\documentclass{article}

\usepackage{amsmath,amsfonts,bm}

\def\eqref#1{equation~\ref{#1}}

\def\1{\bm{1}}

\DeclareMathAlphabet{\mathsfit}{\encodingdefault}{\sfdefault}{m}{sl}
\SetMathAlphabet{\mathsfit}{bold}{\encodingdefault}{\sfdefault}{bx}{n}

\newcommand{\KL}{\mathrm{KL}}

\DeclareMathOperator*{\argmax}{arg\,max}

\usepackage{amsmath}
\usepackage{dsfont}
\usepackage{graphicx}
\usepackage{algorithm}
\usepackage{algorithmic}
\usepackage{nicefrac}
\usepackage[utf8]{inputenc}
\usepackage[english]{babel}
\usepackage{wrapfig}
\usepackage{booktabs}
\usepackage{amsthm}
\usepackage{ amssymb }

\usepackage[final]{corl_2020} 

\newcommand{\J}{\mathcal{J}}
\newcommand{\D}{\mathcal{D}}
\newcommand{\piprior}{\pi^{\text{prior}}}
\newcommand{\expt}[2]{\mathbb{E}_{#1} \left[ #2 \right]}
\newcommand*\diff{\mathop{}\!\mathrm{d}}

\title{Learning Dexterous Manipulation from\\ Suboptimal Experts}

\author{
  Rae Jeong\\
  DeepMind\\
  \texttt{raejeong@google.com} \\
  \And
  Jost Tobias Springenberg\\
  DeepMind\\
  \texttt{springenberg@google.com} \\
  \And
  Jackie Kay\\
  DeepMind\\
  \texttt{kayj@google.com} \\
  \And
  Daniel Zheng\\
  DeepMind\\
  \texttt{dhhzheng@google.com} \\
  \And
  Yuxiang Zhou\\
  DeepMind\\
  \texttt{yuxiangzhou@google.com} \\
  \And
  Alexandre Galashov\\
  DeepMind\\
  \texttt{agalashov@google.com} \\
  \And
  Nicolas Heess\\
  DeepMind\\
  \texttt{heess@google.com} \\
  \And
  Francesco Nori\\
  DeepMind\\
  \texttt{fnori@google.com} \\
}
\begin{document}

\maketitle

\begin{abstract}
Learning dexterous manipulation in high-dimensional state-action spaces is an important open challenge with exploration presenting a major bottleneck. Although in many cases the learning process could be guided by demonstrations or other suboptimal experts, current RL algorithms for continuous action spaces often fail to effectively utilize combinations of highly off-policy expert data and on-policy exploration data. 
As a solution, we introduce \textit{Relative Entropy Q-Learning} (REQ), a simple policy iteration algorithm that combines ideas from successful offline and conventional RL algorithms. It represents the optimal policy via importance sampling from a learned prior and is well-suited to take advantage of mixed data distributions.
We demonstrate experimentally that REQ outperforms several strong baselines on robotic manipulation tasks for which suboptimal experts are available. We show how suboptimal experts can be constructed effectively by composing simple waypoint tracking controllers, and we also show how learned primitives can be combined with waypoint controllers to obtain reference behaviors to bootstrap a complex manipulation task on a simulated bimanual robot with human-like hands. Finally, we show that REQ is also effective for general off-policy RL, offline RL, and RL from demonstrations. Videos and further materials are available at \href{https://sites.google.com/view/rlfse/}{\textcolor{blue}{sites.google.com/view/rlfse}}.
\end{abstract}

\keywords{robotic manipulation, deep reinforcement learning, imitation learning} 


\section{Introduction}
\label{sec:introduction}

	In recent years, deep reinforcement learning (RL) algorithms have demonstrated increasing capabilities in solving complex robotic control problems both in simulation and on real robots. However, exploration remains a significant challenge for high-dimensional robotics tasks with sparse rewards -- the prevalent setting in the domain of robotic manipulation. Crafting a shaped reward function for manipulation tasks is often highly non-trivial due to the characteristics of the desired behavior, which might require complex interactions with tools to accomplish the task.
	Constructing controllers that achieve partial task success and exhibit the desired behavior, on the other hand, is often possible with methods from classical robotics, such as defining waypoint tracking controllers for picking up tools in the environment. 
	Ideally, our RL algorithms should be flexible enough to incorporate off-policy data generated by such a ``suboptimal  expert".
	Indeed, recent work on deep RL from demonstrations (RLfD) \citep{ddpgfd, rajeswaran2017learning} as well as offline RL from fixed datasets \citep{crr,awac} has partially made progress in this direction. However, we find that current algorithms are ineffective in utilizing highly off-policy data generated from a suboptimal expert together with exploration data generated by their own policy.
	
	In this work, we aim to develop an algorithm that excels in the setting of reinforcement learning from suboptimal experts (RLfSE) where we have direct access to an imperfect or partial task solution, as well as the ability to collect new data. To that end, we present three main contributions: 1) we develop a general policy iteration algorithm, \textit{Relative Entropy Q-Learning} (REQ), which takes advantage of highly off-policy exploration data and shows strong performance across off-policy, offline RL, and RLfD; 2) we show that a simple exploration strategy which intertwines the policy's actions with a suboptimal expert's actions results in an effective exploration strategy for complex robotic manipulation tasks; 3) we demonstrate that suboptimal experts can be constructed by composing learned primitives and waypoint tracking controllers, allowing us to learn challenging dexterous manipulation task on a simulated bimanual robot with human-like hands.
	
    \begin{figure}[t]
    \centering
    \includegraphics[width=1.0\linewidth]{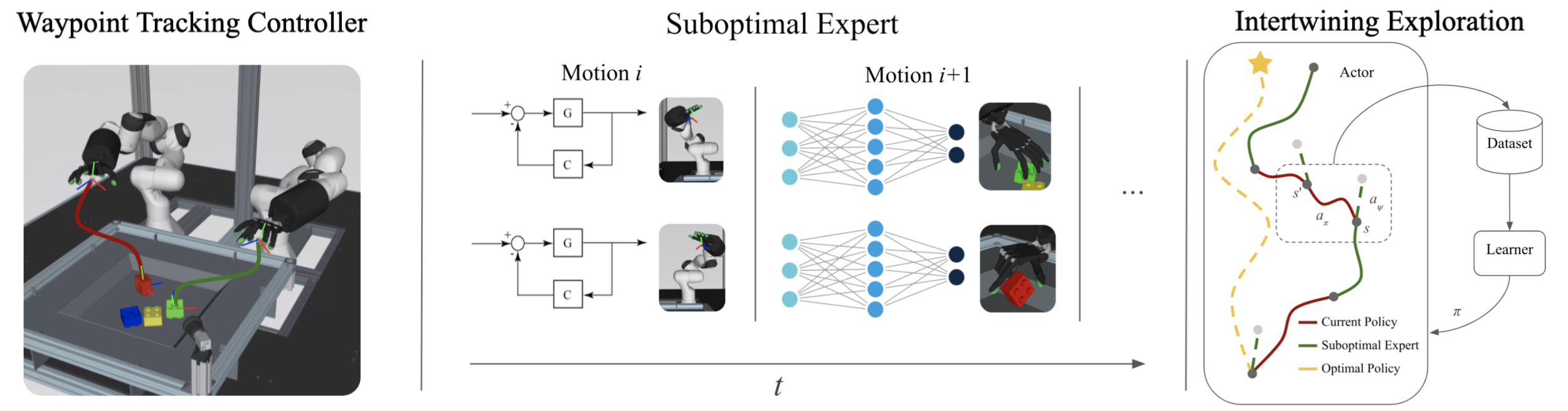}  
    \caption{Left: Bimanual Shadow Hand LEGO stacking task with example waypoints. Middle: Motions are concatenated sequentially to construct a useful suboptimal expert. The robot arm end effector poses are controlled with waypoint tracking controllers, while the dexterous human-like hands are controlled with learned primitives. Right: The suboptimal experts' actions can be intertwined with the current policy's actions to achieve rich exploration. Note that it is also possible to record the actions that the suboptimal expert would have taken, shown as dotted lines.}
    \label{fig:overview}
    \end{figure}

\section{Preliminaries}
\label{sec:prelim}

    We consider the standard reinforcement learning setting in a Markov Decision Process (MDP) described by the tuple $(\mathcal{S}, \mathcal{A}, r, \gamma, P, \mu_0)$ consisting of: the state space $\mathcal{S}$, the action space $\mathcal{A}$, the instantaneous reward function $r(s_t, a_t)$ -- obtained by executing action $a_t$ in state $s_t$ -- the discount $\gamma$, the transition distribution $P(s_{t+1} | s_t, a_t)$ and the initial state distribution $\mu_0$. The RL objective $\J$ corresponds to maximizing the expected reward under a given policy $\J(\pi) = \expt{\pi, P}{{\textstyle\sum}_{t=0}^\infty \gamma^t r(s_t, a_t) | s_0 \sim \mu_0(s)}$, where we have $a_t \sim \pi(\cdot | s_t)$ and $s_{t+1} \sim P(\cdot | s_t, a_t)$. We use $\pi^* = \argmax_{\pi} \J(\pi)$ to denote the optimal policy. We make use of the state-action value function for policy $\pi$, also known as the Q-function which can be defined recursively as $Q^\pi(s_t, a_t) = r(s_t, a_t) + \gamma \expt{\pi, P}{Q^\pi(s_{t+1}, a_{t+1})}$, with which we can express $\J$ in the alternate form $\J(\pi) = \expt{\pi, P}{Q^\pi(s_t, a_t) | s_0 \sim \mu_0(s)}$. Furthermore, we consider a policy iteration scheme \citep{sutton_rl_book} with additional KL constraints \citep{reps, mpo, rerpi}. In this setting the goal is to maximize the Q-values while staying close to a reference (or prior) policy $\piprior$. In particular, given a current policy $\pi_{i-1}$ that we want to improve upon, we can write the objective for iteration $i$ as a constrained optimization problem, assuming access to a dataset $\mathcal{D}$:
    \vspace*{-0.15cm}
    \begin{equation}
    \begin{aligned}
        \pi_i = \arg \max_\pi \J_{c}(\pi, \piprior, \epsilon) & = \arg \max_\pi \expt{s\sim \D}{ \expt{a \sim \pi(\cdot | s)}{Q^{\pi_{i-1}}(a, s)}} \\
        & \qquad \textrm{s.t.} \quad \forall s \in \D: \KL(\pi(\cdot | s) \| \piprior(\cdot | s)) \leq \epsilon,
        \label{eq:constrained_kl} \\
    \end{aligned}
    \end{equation}
    where the policy to be optimized is denoted as $\pi(a | s)$, $\piprior$ is the prior policy, and $\KL$ denotes the Kullback-Leibler divergence. Note that from the perspective presented here, $\piprior$ is not necessarily a guess for an optimal policy but serves as the prior towards which we regularize.\footnote{Note that the per-state KL constraint is often relaxed to an average constraint over states from the buffer; e.g. in \citet{awac, mpo, abm}.} Successively solving the optimization from \autoref{eq:constrained_kl} for $i \in [1, N]$ -- with interleaved policy evaluation to obtain $Q^\pi$ -- then constitutes the policy iteration loop.


\section{Relative Entropy Q-Learning}
\label{sec:req}

    We introduce \textit{Relative Entropy Q-Learning} (REQ). REQ is a policy iteration algorithm targeting the KL-constrained RL objective $\J_c$ from \autoref{eq:constrained_kl} in each iteration. 
    We start by realizing that the solution to the KL-constrained objective at iteration i, $\pi_{i} = \arg \max_\pi \J_{c}(\pi, \piprior_i, \epsilon)$, can be obtained in closed form by formulating the Lagrangian of the constrained optimization problem and solving for $\pi$. The solution consists of a softmax over Q-values (a well known result, see e.g. \citep{reps, mpo, rerpi, abm, rawlik}) $\pi_i(a|s) \propto \piprior_i(a|s) \exp{(Q^{\pi_{i-1}}(s,a)/\eta_s})$ where the temperature $\eta_s$ can be obtained by solving the dual function of $\J_{c}$; a convex optimization problem. We refer to the Appendix \ref{sec:reps_derivation} for a derivation of the Lagrangian as well as how we optimize the dual function for $\eta_s$. Exactly calculating the normalization constant is intractable in continuous action spaces but we can, however, sample from $\pi_i(a | s)$ via importance weighting of samples from $\piprior_i$ -- an observation that we will now use to define the REQ policy evaluation step.
    
    \paragraph{\textbf{Policy Evaluation}} The first key observation is that we can learn the state-action value function of $\pi_i$ without the need to explicitly represent $\pi_i$ via a parametric policy. Instead, we can realize that the squared temporal difference error $(Q^{\pi_i}(s_t, a_t) - ( r(s_t, a_t) + \gamma \expt{\pi_i}{Q^{\pi_i}(s_{t+1}, a_{t+1})}))^2$ can be evaluated using importance sampling, leading to the following objective:

    \vspace*{-0.6cm}
    \begin{equation}
        Q^{\pi_i}_\phi = \arg \min_{Q_\phi} \mathop{\mathbb{E}}_{s, a, r, s' \in \D} \Big[\Big( Q_\phi(s, a) - \Big[r + \gamma \sum_{j=1}^N \frac{\exp(Q_{\phi'}(s', a^j)/\eta_{s'})}{\sum_k \exp(Q_{\phi'}(s', a^k)/\eta_{s'})} Q_{\phi'}(s', a^j)\Big]\Big)^2\Big],
    \label{eq:iw_req}
    \end{equation}
    with $\forall j: a^j \sim \piprior_i(\cdot | s')$, where $\phi'$ denotes the parameters of a target network \citep{mnih2015humanlevel} and we estimate the expectation $\expt{\pi _i}{Q_\phi(s', a)} \propto \expt{a' \sim \piprior(\cdot | s')}{\exp(Q_\phi(s', a')/\eta_{s'}) Q_\phi(s', a)}$ with self-normalized importance sampling based on $N$ samples.
    
    A few observations can be made about this objective. The learned Q-function corresponds to the one considered in ABM+MPO \citep{abm}, but with the difference that $\pi_i$ is never projected onto a parametric policy. Instead it is only implicitly represented via importance sampling from the prior -- we thus only need to learn $\piprior$ and a Q-function. This can be beneficial when the prior is learned from data but is not well aligned with high-value regions in $Q^{\pi_i}$. In such a case $\pi_i$ may become multimodal and hard to project to a parametric policy without accumulation of errors. Note that REQ can still represent the optimal policy, as for $\epsilon \rightarrow \infty$ the policy $\pi_i$ will approach $\pi^*$. Hence, the constraint $\epsilon$ allows us to trade-off the exploitation of the Q-function and regularizing towards $\piprior$. An alternative view of the REQ policy evaluation is to consider it as a policy iteration algorithm that uses the the softmax Bellman operator \citep{softmax_operator, littman_thesis} for policy evaluation (with an adaptive method to satisfy a hard KL constraint with respect to a given prior). Further analysis of the REQ policy evaluation is provided in the Appendix \ref{sec:req_theory}.
    
    \paragraph{\textbf{Prior Policy Improvement}} The Q-learning like algorithm from above can work with any prior as long as $\piprior$ has probability mass everywhere; $\piprior(a | s) > 0 \enskip \forall a$. However, the sample based importance weighting scheme from \autoref{eq:iw_req} becomes ineffective -- potentially leading to learning Q-values of a suboptimal policy -- if the number of samples $N$ is low and the prior has small probability mass at actions with high Q-values. The policy improvement step of REQ thus is to learn an effective prior, $\piprior_i$, and improve it in each iteration. We achieve this by fitting the prior to all actions from the dataset, $\D$, whose value is estimated to be higher than the average value of the policy. 
    Formally, we find
    \begin{equation}
    \label{eq:policy_improvement}
        \piprior_{i+1} = \arg \max_{\piprior_\theta} \expt{a, s \sim \D}{ 1\Big[ Q_{\phi'}^{\pi_i}(s, a) \geq \mathbb{E}_{a \sim \pi_i(\cdot | s)}[Q_{\phi'}^{\pi_i}(s, a)] \Big] \log \piprior_\theta(a | s)},
    \end{equation}
    where 1 is the indicator function. That is, we consider learning a prior similar to recent offline RL algorithms such as ABM \citep{abm} and CRR-bin \citep{crr}.
    To avoid overfitting during training (e.g. due to suboptimal Q-values shrinking the prior distribution) we additionally regularize the prior update step. In particular, we employ constraints on the movement of the prior-policy mean ($\KL(\piprior_{i+1}(a|s) \,\|\, \piprior_i(a|s;\Sigma=\Sigma_i) < \epsilon_{\Sigma}$) and the covariance ($\KL(\piprior_{i+1}(a|s_j) \,\|\, \piprior_i(a | s;\mu=\mu_i) < \epsilon_{\mu}$) for a Gaussian prior -- analogous to the trust-regions used for policy optimization in MPO \citep{mpo}, which we can enforce via a simple Lagrangian relaxation approach similar to MPO \citep{rerpi}.
    
    \paragraph{\textbf{Practical algorithm}}
    A full listing of our procedure is presented in Algorithm \ref{alg:req}. In contrast to previous works \citep{mpo}, we formulate a per-state KL constraint $\eta_s$ that we optimize for each state in the batch; i.e. we perform multiple gradient steps on the dual for a given $\eta_s$ to ensure the constraint is tight. In addition, instead of fully optimizing the policy and Q-function in each iteration, we switch to a new iteration after a fixed amount of gradient descent steps (via the use of target networks). 

    \begin{algorithm}[H]
    \small
    \caption{Relative Entropy Q-learning (REQ)}\label{alg:req}
    \begin{algorithmic}
    \STATE \textbf{Input:} number of learning steps $N$, steps between target updates $U$, number of action samples $M$, KL regularization parameter $\epsilon$, initial parameters for $\theta, \phi$ and $\eta_s$ 
    \STATE \textbf{def REQ\_update($\theta, \pi, \mathcal{B}$):}
    \STATE \quad // For this step let $\pi(a | s) \propto \pi_{\theta'}(a | s) \exp(Q_{\phi'}(a, s) / \eta_s)$ and $A^\pi(a, s) = Q_{\phi'}(a, s) - V^\pi(a, s)$
    \STATE \quad Find $\eta_s$ for $s \in \mathcal{B}$ via gradient: $\nabla_{\eta_s} \frac{1}{|\mathcal{B}|} \eta_s \epsilon+\eta_s \log \frac{1}{M} \sum_{j=1}^M \big[ \exp\left(\nicefrac{Q_{\phi'}(s,a_j)}{\eta_s}\right) | a_j \sim \pi_{\theta'}(\cdot, s) \big]$ 
    \STATE \quad Compute $V^\pi(s) = \sum_{j=1}^M \frac{\exp(Q_{\phi'}(a_j, s)/\eta_s)}{\sum_{j=1}^M \exp(Q_{\phi'}(a_j, s)/\eta_s)} Q_{\phi'}(a_j, s),$ where $a_j \sim \piprior_{\theta'}(\cdot | s)$ 
    \STATE \quad Update Q-function with gradient: $\nabla_\phi \frac{1}{|\mathcal{B}|} \sum_{s, a, r, s' \in \mathcal{B}} \Big(r + \gamma V^\pi(s') - Q_\phi(a, s)\Big)^2$
    \STATE \quad Update prior $\pi_\theta$ with gradient: $\nabla_\theta -\frac{1}{|\mathcal{B}|} \sum_{s, a, r \in \mathcal{B}} 1[ A^\pi(a, s) \geq 0\big] \log \pi_\theta(a |s)$
    \STATE \textbf{Initialize:} $N = 0$, $\theta' = \theta$, $\phi' = \phi$
    \WHILE{$i \leq N$}
    \STATE \textbf{Optionally:} collect new data $\D_i$ by following $\pi_i$ or some mixture of $\pi_i$ and an expert policy $\psi$
    \STATE Let $\D \leftarrow \D \bigcup \D_i$
    \STATE sample a batch $\mathcal{B}$ from replay buffer $\D$ 
    \STATE execute \textbf{REQ\_update($\theta, \pi, \mathcal{B}$)}
    \STATE Update policy and Q-function every $U$ steps by copying: $\theta' \leftarrow \theta$, $\phi' \leftarrow \phi$
    \ENDWHILE
    \end{algorithmic}
    \end{algorithm}


\section{Reinforcement Learning from Suboptimal Experts}
\label{sec:rlfse}

    In this section we explain how our method can be used for reinforcement learning from suboptimal experts and describe a class of suboptimal experts for robotic manipulation consisting of simple waypoint tracking controllers.
    
    \paragraph{\textbf{Problem Formulation}} We consider an RL setting with additional access to a suboptimal expert $\psi(a|s)$. We assume that $\psi$ exhibits behaviors that are relevant to the task but is not necessarily the optimal policy $\pi^*$. We refer to this setting as \textit{Reinforcement Learning from Suboptimal Experts} (RLfSE). In RLfSE, our policy iteration scheme can be understood as a form of sample-based approximate policy iteration from mixed behavior data -- somewhat similar to the AggreVaTe family of algorithms \citep{aggrevate,aggrevated}. This setting is motivated by real world problems for which domain-specific solutions are already deployed. RLfSE can give us access to broader data distributions than reinforcement learning from demonstrations (RLfD \citep{ddpgfd, rajeswaran2017learning}), as we can choose to collect data from $\psi$ or a mixture of $\psi$ and $\pi$. Additionally, direct access to $\psi$ also allows us to label off-policy data with the expert's actions, similar to commonly used no-regret imitation learning algorithms \citep{aggrevate, dagger,lols}. 

	\paragraph{\textbf{Waypoint Tracking Controllers}} As a concrete example, for each manipulation task, we construct a suboptimal expert $\psi(a | s)$ by composing waypoint (pose) tracking controllers. (Note $\psi$ is deterministic in our case, thus we can write $a = \psi(s)$.) Such pose tracking controllers can be formulated by leveraging differential kinematics as well as velocity control modes of robotic arms. Using relative reference frames, these controllers can generalize under homogeneous transformations and provide an intuitive interface for humans to specify waypoints to follow. The pose controllers that we use are linear feedback controllers on the end-effector(s) of the robot arm(s) using velocity control. Formally we use $\psi(s) = [\psi_p(s),  \psi_o(s)]$ for each controllable six degree of freedom with $\psi_p(s) = K_p e_p(s)$ and $\psi_o(s) = K_o e_o(s)$ where $K_p$ and $K_o$ are positive definite gain matrices. The position error is $e_p(s) = p_d(s) - p_t(s),$ where $p_t$ and $p_d$ are the measured and desired positions of the end-effector, respectively. We define the \textit{orientation error} using unit quaternions where $\mathcal{Q}_d = \{\eta_d, \epsilon_d\}$ and $\mathcal{Q}_t = \{\eta_t, \epsilon_t\}$ represent the desired and measured orientations respectively, with $\eta$ representing the real valued quaternion components and $\epsilon$ the imaginary values. We define the orientation error as $e_o(s) = \Delta\epsilon = \eta_t(s)\epsilon_d(s) - \eta_d(s)\epsilon_t(s) - \textbf{S}(\epsilon_d(s))\epsilon_t(s)$ where $\textbf{S}(\cdot)$ is the skew-symmetric operator \citep{siciliano_robotics_book}. Additional details of the composition of the waypoint tracking controller are presented in Appendix \ref{sec:internal_agent}.
	
	\paragraph{\textbf{Relative Entropy Q-learning from Suboptimal Experts}} To make use of the suboptimal expert $\psi$, we propose a simple exploration strategy: intertwining the execution of the current policy $\pi$ with the execution of $\psi$. For each episode we first randomly choose with probability $\lambda_{intertwine}$ whether to execute a mix of the policy and the expert or either of the policy or the expert only. In episodes in which we choose actions according to a mixture of policy and expert we execute expert's action with probability $\lambda_{\psi}$ at every time step. Otherwise, we execute the entire episode with probability $\lambda_{\psi}$. This is illustrated in \autoref{fig:overview} and the full procedure is presented in Appendix \ref{sec:intertwining} Algorithm \autoref{alg:intertwining}. Note that setting $\lambda_{intertwine} = 0$ recovers the RLfD setting, where a certain portion of the data in the replay buffer are demonstrations from the suboptimal expert $\psi$. In addition to using $\psi$ for data generation, we can also take advantage of access to the expert in the prior policy improvement step of REQ by using the following equation,
    \begin{equation}
    \begin{aligned}
    \label{eq:reqfse_policy_improvement}
        \piprior_{i+1} = \arg \max_{\pi_\theta} \expt{a, s \sim \D}{\big(1[ A^q(a, s) \geq 0\big]  + \expt{\Bar{a} \sim \psi}{1[ A^q(\bar{a}, s) \geq 0\big]}\big) \log \pi_\theta(a |s)},\\
    \end{aligned}
    \end{equation}
    where we now consider actions sampled from $\pi$ and actions from the expert $\psi$ for inclusion into the prior, by evaluating the suboptimal expert on states from $\D$. Note that since $\psi$ is deterministic in our case, i.e. the expectation is over a delta distribution and can be evaluated using the single $\bar{a} = \psi(s)$.
    
    
\section{Related Work}
\label{sec:citations}

	\textbf{Reinforcement Learning from Demonstrations} Our approach can be seen as a form of imitation learning from imperfect experts. Similar to DDPGfD \cite{ddpgfd}, DQNfD \cite{dqnfd} and MPOfD \cite{mpofd} we use off-policy data generated by experts in the replay buffer to augment reinforcement learning. However, our reinforcement learning from suboptimal experts \citep{acteach, residual_rl} setup allows us to both gather a mixture of data from the current policy and the suboptimal expert, as well as label any states with what the suboptimal expert would have done. Querying expert actions in states visited by the policy bears similarity to DAgger \cite{dagger}. \\
	\textbf{Off-Policy Reinforcement Learning} When no experts are used and all data is collected by following the policy $\pi$, 
	our algorithm bears similarity to a number of recent actor-critic and Q-learning algorithms. In the limit of $\epsilon \rightarrow \infty$ it can be seen as a form of deep Q-learning \citep{mnih2015humanlevel, qlearning} where the maximization of the Q-function is performed approximately via an amortized proposal -- an idea that has previously been considered in the literature using a mixture of a uniform and a learned proposal fitted to the best current action for each state \citep{vandewiele2020qlearning}. Our method can also directly be related to actor-critic algorithms that implement constrained policy updates -- which have seen resurgence in recent years \citep{reps, mpo, rerpi, trpo, ppo}.
	In particular REQ can be compared to MPO \citep{mpo} which optimizes the objective from \autoref{eq:constrained_kl}, but uses an expected constraint and chooses the prior to be the policy $\pi_{i-1}$ from the last iteration -- i.e. it performs an EM-style policy optimization with a trust-region constraint. In addition MPO performs a projection step to obtain a parametric representation of $\pi_i$ in each iteration, whereas REQ only learns the prior and represents the policy via importance sampling. \\
	\textbf{KL Regularized Reinforcement Learning} Whereas REQ performs constrained updates with respect to the maximum reward objective, many recent algorithms directly optimize a regularized objective that trades off reward with the divergence to a fixed \cite{hausman2018learning,sac,haarnoja2017reinforcement,hunt2019composing} or learned \cite{distral,galashov2018information,tirumala2019exploiting} reference distribution. When the reference distribution is learned it plays a similar role to our prior. Learned reference distributions have so far mainly been considered for transfer and multi-task learning \citep{distral,galashov2018information,tirumala2019exploiting,iwpa}.\\
	\textbf{Offline Reinforcement Learning} Our algorithm can also be seen as a simplification of a recent offline RL algorithm, ABM+MPO \citep{abm} -- where we only learn the prior and represent the optimal policy implicitly. While this may seem limiting, we find that removing the need for two policies -- which also makes our algorithm closely related to \citep{crr} -- results in better performance in practice. A similar effect was also observed in \citet{crr,awac} where the authors found that CRR/AWAC outperformed ABM in high-dimensions. Note that the main difference between REQ and CRR is that we perform policy evaluation on the constrained optimal policy and treat the learned policy as the prior -- CRR instead considers the behavior distribution (the dataset) as the prior. As the experiments will show, this allows us to obtain significantly better learning speed when new data can be collected.


    \begin{figure}[t]
    \centering
    \includegraphics[width=1.0\linewidth]{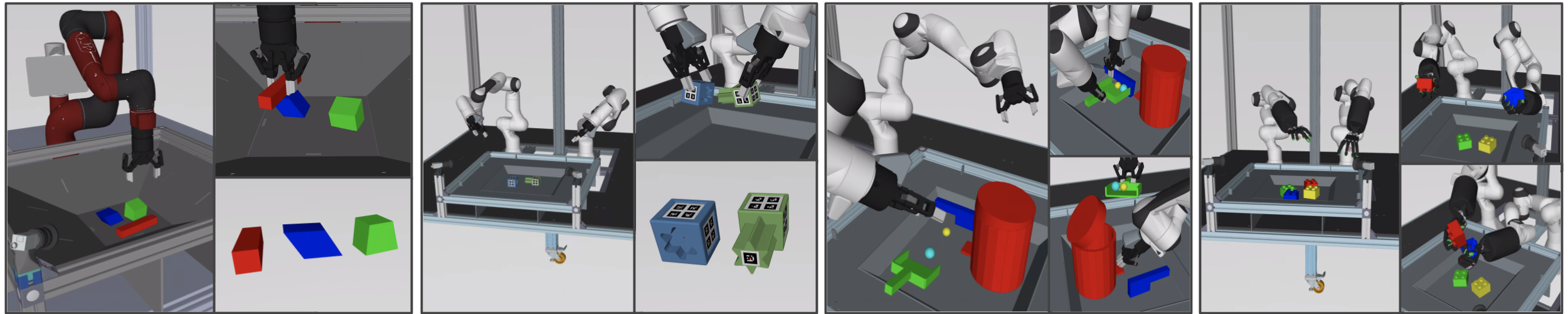}  
    \caption{Simulated manipulation environments in the MuJoCo simulator \citep{mujoco}. From left to right: single arm stacking, bimanual insertion, bimanual cleanup and bimanual Shadow hand LEGO assembly. The environments are described in greater detail in Appendix \ref{sec:environments}.}
    \label{fig:rlfse_envs}
    \end{figure}
    
\section{Experimental Results}
\label{sec:result}

    We evaluate our algorithms, REQ and REQfSE, in several different reinforcement learning settings and demonstrate that the same algorithm performs well in all regimes. First, in section \ref{sec:experiments:rlfd} we consider reinforcement learning from demonstrations (RLfD) and reinforcement learning from suboptimal experts (RLfSE) for the manipulation tasks shown in \autoref{fig:rlfse_envs}. Next, in section \ref{sec:experiments:rl}, we evaluate REQ for (off-policy) reinforcement learning (on the DeepMind Control Suite benchmark \citep{dm_control}) as well as in a pure offline reinforcement learning setting. Finally, we demonstrate REQfSE's potential to solve challenging dexterous manipulation tasks by controlling a simulated bimanual robot with human-like hands. Here, we construct a suboptimal expert by composing waypoint controllers with learned primitives (trained via REQ) which we then employ in the context of REQfSE. Hyperparameters for all algorithms can be found in the Appendix \ref{sec:hyper}.
    
	\subsection{Reinforcement Learning from Demonstrations and Suboptimal Experts}
	\label{sec:experiments:rlfd}
\begin{figure}[t]
    \centering
    \includegraphics[width=1.0\linewidth]{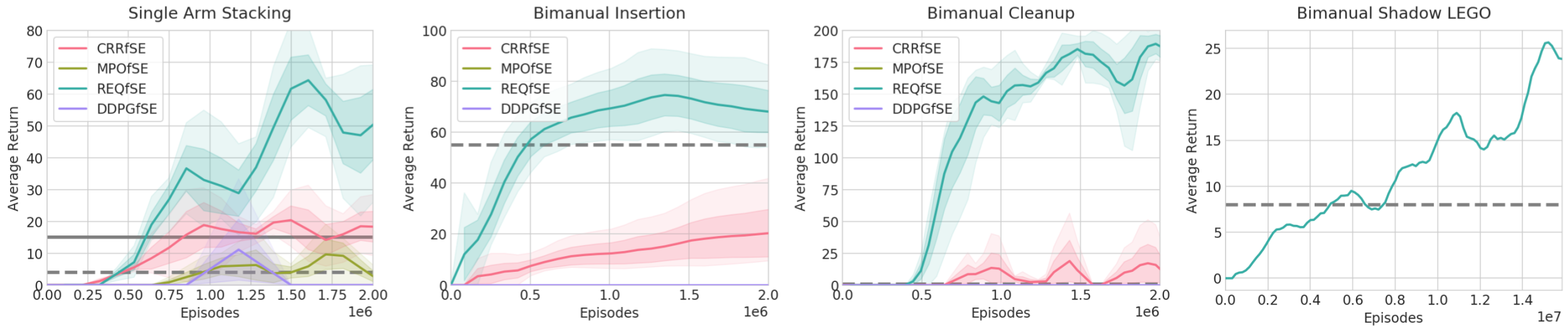}  
    \caption{
    RLfD and RLfSE results on the tasks shown in \autoref{fig:rlfse_envs}. The dotted line represents the performance of the suboptimal expert and the solid line shows the performance of the best RLfD agent over CRRfD, MPOfD, REQfD and DDPGfD. In both the bimanual insertion and the cleanup tasks the performance of RLfD agents is zero. For the bimanual Shadow LEGO task, we only evaluate REQfSE as this task is significantly harder than the first three. Additional comparisons of RLfSE, RLfD, BC, DAgger \citep{dagger} and Residual RL \citep{residual_rl} are available in the Appendix \ref{sec:rlfd_rlfse}.     
    }
    \label{fig:rlfse_rlfd_plot}
    \end{figure}

	REQ bears similarity to both recent offline RL algorithms as well as to conventional policy iteration algorithms such as MPO. We thus hypothesize that REQ is well suited to take advantage of highly off-policy data but may also be able to make effective use of near on-policy data.
    We compare REQ to existing algorithms for learning from demonstrations in complex manipulation environments in both the RLfD and RLfSE settings. Specifically, we compare REQ to DDPGfD \citep{ddpgfd, s2r_deform} and MPOfD \citep{mpofd}, which both make use of demonstration data that is added to the replay buffer. Additionally, we perform an ablation on the importance of the REQ policy evaluation by replacing the importance weighted formulation from \autoref{eq:iw_req} with the standard TD(0) operator \citep{sutton_rl_book} which effectively recovers CRR-bin \citep{crr}, a strong offline RL baseline, but with  a trust-region in the policy improvement and without distributional Q-learning.\footnote{Note also that CRR is equivalent to AWAC \citep{awac} when exponential weighting rather than binary weighting is used, a setting that we found to not perform better in our experiments.} We refer to this as CRRfSE in our experiments. \autoref{fig:rlfse_rlfd_plot} shows that especially in the bimanual environments, training with REQ provides a significant performance gain compared to other algorithms. We speculate that as the state-action space grows, using highly off-policy data for policy evaluation becomes significantly more challenging, similar to the difficulties encountered in offline RL with high state-action spaces. Given that most off-policy RL algorithms -- unlike REQ or CRR -- are designed to consider only the actions from their respective policy $\pi$ for policy improvement, ignoring potentially good actions in the dataset can make them significantly less effective. This is evident in our results where CRR performs well compared to DDPG and MPO but significantly underperforms REQ. We highlight that CRR's difference in speed can be solely attributed to the policy evaluation step: CRR performs policy evaluation with respect to what we consider to be the prior policy, but REQ can sharpen the prior to obtain better Q-values (and in turn a better policy), greatly improving learning speed. 
    
    RLfSE also significantly outperforms RLfD. We speculate that at the beginning of learning, there may be a large difference in the state distribution between the demonstrations and on-policy experience. This can make it difficult to exploit the expert actions for policy evaluation and policy improvement. Executing suboptimal experts from states visited by the policy (as happens during ``intertwining") makes expert's action available for states close to the policy's state distribution, thus rendering the data more effective. Intertwining also has the effect of broadening the state-visitation distribution, allowing for a more robust value learning. Furthermore, as expected, REQfSE significantly outperforms the suboptimal experts alone.
    Additional ablations of the REQ algorithm can be found in Appendix \ref{sec:ablation}. Finally, it is also worth observing the resulting behaviors of the policies optimized with REQfSE (footage available in the supplementary video). For example, the bimanual cleanup task shown in \autoref{fig:rlfse_envs} features two robotic arms in a workspace containing a trashcan, brush, dustpan and some balls. The goal is to pick up both the brush and the dustpan to sweep up the balls and place it in the trashcan. We find that when using a suboptimal expert with a low success rate to guide exploration (via intertwining) this difficult task can be learned successfully using a sparse reward that only indicates task success but provides no additional information, e.g. tool use. This suggests that REQfSE is well-suited to tasks where humans can provide additional cues for desired behavior or intermediary solution steps that are difficult to capture with a reward function.

	\subsection{Off-Policy and Offline Reinforcement Learning}
	\label{sec:experiments:rl}
	
	\begin{figure}[b]
    \centering
    \includegraphics[width=1.0\linewidth]{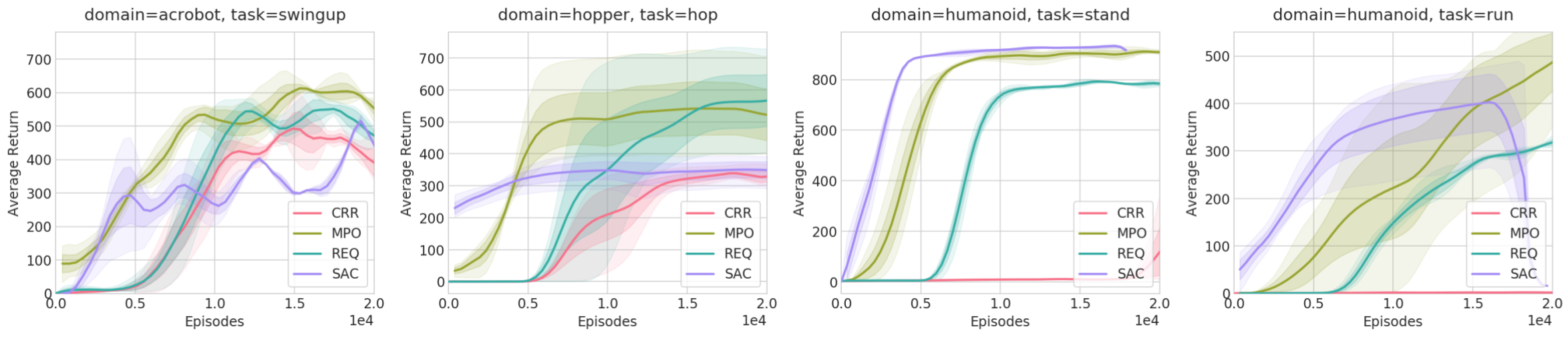}  
    \caption{Online off-policy RL results on the DeepMind Control Suite \citep{dm_control}. MPO and SAC results matches \citep{acme} and \citep{pytorch_sac} respectively. Note that SAC sometimes becomes unstable when run for longer.}
    \label{fig:control_suite_benchmark}
    \end{figure}
    
    To better understand the REQ algorithm, we evaluate it in the off-policy and offline RL settings. These settings can be interpreted as two ends of a spectrum: one end with access to data generated solely by the optimized policy, and the other where only offline data is available, with RLfSE and RLfD somewhere in the middle. Below we compare REQ to specialized algorithms for each setting. We hypothesize that in the standard off-policy RL setting REQ can perform significantly better than the closely related offline RL algorithm CRR. Recall that CRR considers the data itself to be the prior and hence performs policy evaluation for what we refer to as $\piprior$. In contrast REQ uses the optimal KL-constrained policy in each iteration to estimate the value (\autoref{eq:iw_req}). As a result, we would expect CRR to learn slowly when the data is bad or random. In contrast, REQ's importance weighted policy can filter out the good actions for each state (subject to the KL constraint) and should result in evaluation of a better policy and thus faster learning. To test this hypothesis, we evaluate REQ and CRR as well as two state of the art off-policy RL algorithms (MPO \citep{mpo} and SAC \citep{sac}) on the DeepMind Control Suite \citep{dm_control}. We again use the variant of CRR-bin \citep{crr} which differs from REQ only in the policy evaluation step. The results are shown in \autoref{fig:control_suite_benchmark}. REQ's policy evaluation indeed results in a significant improvement upon CRR. REQ's performance is also comparable to state of the art off-policy RL algorithms, although it learns more slowly for the humanoid environments.
    
    We further hypothesize that REQ would also be competitive when we consider training from a fixed offline dataset. To this end, we compare REQ to the results of \citet{crr} for the Control Suite tasks, using the same network architecture and evaluation criteria. As can be seen in \autoref{tbl:offline}, REQ performs comparatively to CRR when the same advantage transformation is used (CRR-bin). Interestingly, the improved policy evaluation has less impact in this setting. We hypothesize that this is due to the fact that we need to stay close to the actions contained in the data to avoid problems with Q-value overestimation \citep{crr, abm}. When using another variant, REQ-exp, which uses the same exponential advantage transformation as CRR-exp, we again observe comparable performance. Overall, the results in off-policy and offline RL highlight the generality of the REQ algorithm. Although it does not completely outperform specialized algorithms for either setting, it simultaneously performs well in both. It also sheds light on important algorithmic components for learning efficiently and learning from highly off-policy data. The former is achieved by considering a KL-constrained optimal policy in the algorithm, like many of the algorithms for off-policy RL \citep{mpo, rerpi}. The latter is achieved through considering the actions from the dataset, rather than the actions from the policy being optimized as proposed in recent offline RL algorithms \citep{crr, abm}.
    
    \begin{table}[t]
    \small
    \begin{center}
    \begin{tabular}{lrrrrr|rr} \toprule
    {} & {BC} & {D4PG} & {ABM} & {CRR-exp} & {CRR-bin} & {REQ} & {REQ-exp} \\ \midrule
    Cartpole  & $386 \pm 6$ & $855 \pm 13$ & $798 \pm 30$ & $664 \pm 22$ & $\mathbf{860 \pm 7}$ & $855 \pm 10$ & $741 \pm 29$  \\
    Finger Turn  & $261 \pm 39$ & $\mathbf{764 \pm 24}$ & $566 \pm 25$ & $714 \pm 38$ & $755 \pm 31$ & $720 \pm 41$ & $705 \pm 47$ \\
    Walker Stand  & $386 \pm 6$ & $\mathbf{929 \pm 46}$ & $689 \pm 13$ & $797 \pm 30$ & $881 \pm 13$ & $901 \pm 27$ & $810 \pm 17$ \\
    Walker Walk  & $417 \pm 33$ & $\mathbf{939 \pm 19}$ & $846 \pm 15$ & $901 \pm 12$ & $936 \pm 3$ & $928 \pm 9$ & $912 \pm 13$ \\
    Cheetah Run & $407 \pm 56$ & $308 \pm 121$ & $304 \pm 32$ & $\mathbf{577 \pm 79}$ & $453 \pm 20$ & $438 \pm 21$ & $521 \pm 48$ \\
    Fish Swim  & $466 \pm 8$ & $281 \pm 77$ & $527 \pm 19$ & $517 \pm 21$ & $585 \pm 23$ & $\mathbf{592 \pm 41}$ & $586 \pm 36$  \\ \midrule
    Insert Ball  & $385 \pm 12$ & $154 \pm 54$ & $409 \pm 4$ & $625 \pm 24$ & $\mathbf{654 \pm 42}$ & $638 \pm 54$ & $602 \pm 22$ \\
    Insert Peg  & $324 \pm 31$ & $71 \pm 2$ & $345 \pm 12$ & $387 \pm 36$ & $365 \pm 28$ & $\mathbf{397 \pm 41}$ & $389 \pm 21$ \\
    Humanoid Run  & $382 \pm 2$ & $1 \pm 1$ & $302 \pm 6$ & $586 \pm 6$ & $412 \pm 10$ & $408 \pm 24$ &  $\mathbf{596 \pm 15}$ \\ \bottomrule
    \end{tabular}
    \end{center}
    \caption{Offline RL results on the DeepMind Control Suite \citep{dm_control}. The reference numbers for baselines are from \citet{crr}.}
    \label{tbl:offline}
    \vspace{-0.7cm}
    \end{table}
	
    \subsection{Reinforcement Learning from Composition of Learned Primitives and Controllers}
    
    \label{sec:experiments:composition}
    
    Finally, to demonstrate REQ’s potential on challenging manipulation tasks, we evaluate REQfSE on the bimanual Shadow LEGO task. The goal in this task is to stack two LEGO blocks using a bimanual arm setup with two Shadow Hands. This problem is extremely challenging not only due to its high-dimensional state-action space (state size of 176 and action size of 52), but also because simple waypoint tracking controllers are not sufficient to provide reference behavior for the full task. Therefore, we first learn two primitive policies using REQ in an off-policy RL setting. The first primitive is trained to orient the green LEGO to an upwards facing position (while in the basket) with the left hand. The second primitive is trained to orient the red LEGO block to a downwards facing position with the right hand. Both primitives are trained ``from-scratch" using a shaped reward function as described in the appendix \ref{sec:environments}. We then construct a suboptimal ``expert'' for this task by composing the waypoint tracking controllers for reaching the blocks with the learned primitives to orient the blocks, followed by a third phase in which the LEGO blocks are joined together (again using simple waypoint tracking controllers). The results can be seen in \autoref{fig:rlfse_rlfd_plot} where REQfSE is able to achieve higher performance than the suboptimal expert (footage available in the supplementary video).


\section{Conclusion and Future Work}
\label{sec:conclusion}

	We have presented an approach for learning from suboptimal experts for complex dexterous robotic manipulation. We demonstrated that our algorithm, \textit{Relative Entropy Q-Learning} (REQ), is effective in many different learning scenarios -- including off-policy and offline RL as well as RL from demonstration (RLfD). In addition, our proposed approach for \textit{Reinforcement Learning from Suboptimal Experts} (RLfSE), significantly outperforms competitive baselines. In particular, REQfSE leverages intertwining exploration to solve highly complex tasks that are otherwise intractable. To achieve this, we proposed using waypoint tracking controllers as the suboptimal experts. We argue that this approach can provide an intuitive and simple interface for humans to specify the desired behaviors for the task without requiring human demonstrations or engineering shaped reward functions. 



\clearpage
\acknowledgments{We would like to thank Abbas Abdolmaleki, Akhil Raju, Antoine Laurens, Charles Game, Celine Smith, Christopher Schuster, Claudio Fantacci, Dave Barker, David Khosid, Emre Karagozler, Federico Casarini, Francesco Romano, Giulia Vezzani, Jon Scholz, Jose Enrique Chen, Michael Neunert, Murilo Martins, Nathan Batchelor, Nicole Hurley, Nimrod Gileadi, Nylda Adelise, Oleg Sushkov, Stefano Saliceti, Thomas Lampe, Thomas Rothörl, Tom Erez, Tuomas Haarnoja and Yuval Tassa for fruitful discussion and support.}


\bibliography{example}  

\newpage
\appendix


\section{Appendix}

\subsection{Environments}
\label{sec:environments}
    In this section, we will provide the details of the simulated environments used in our experiments. All of our simulated environments are modelled using the MuJoCo simulator \cite{mujoco}. 
    
    \begin{wrapfigure}{L}{0.4\textwidth}
    \vspace{-0.5cm}
    \centering
    \includegraphics[width=0.4\textwidth]{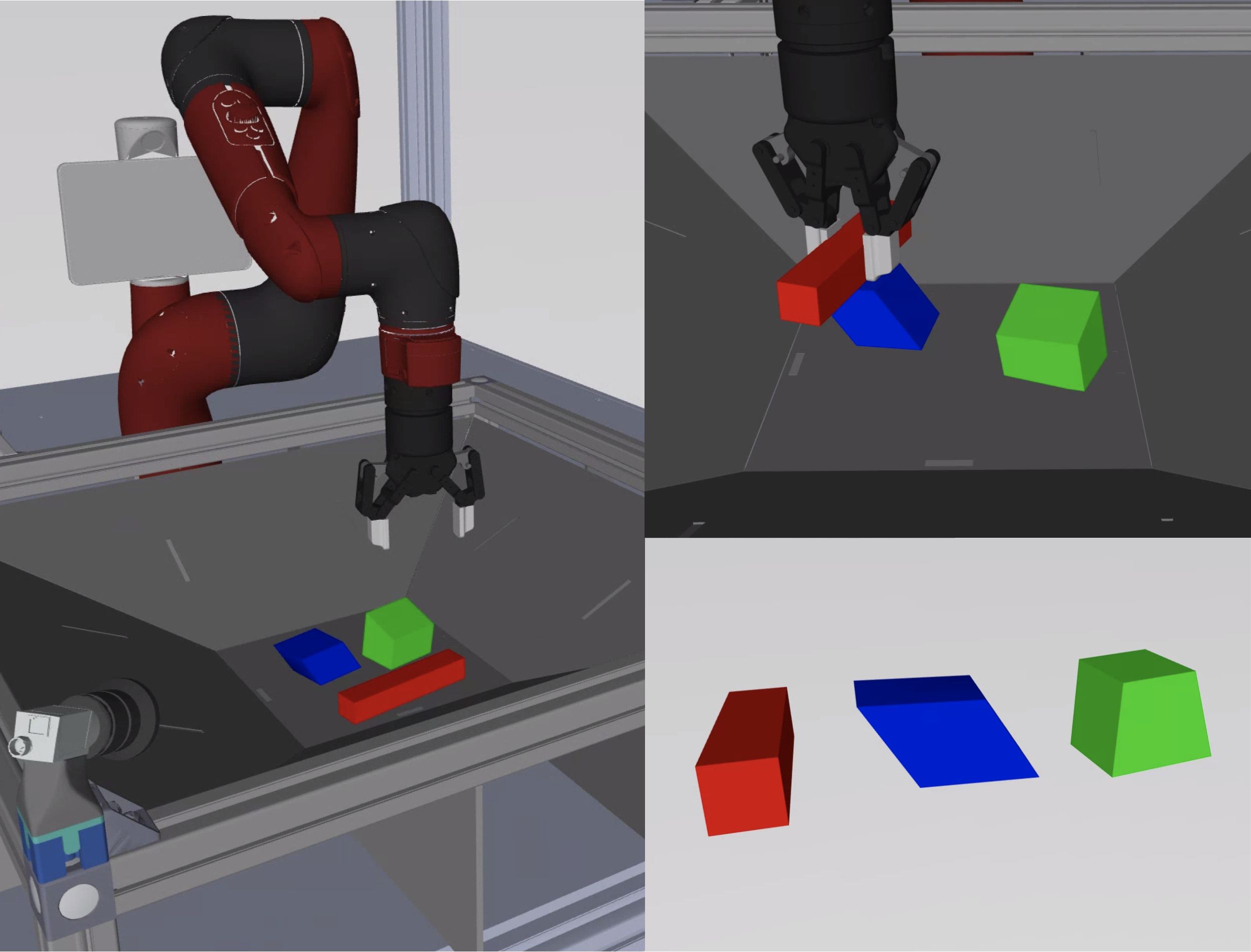}
    \vspace{-1.3cm}
    \end{wrapfigure}
    
    \paragraph{\textbf{Single Arm Stacking (State Size: 69, Action Size: 5)}}This environment features a 7 DOF arm that is controlled with a Cartesian velocity controller with the orientation of the end-effector always pointing down, reducing the action size to translation in X, Y, Z, rotation about the Z axis and the gripper control. The goal of the task is to stack the red object on top of the blue object. The reward is sparse and active when the red object is on the blue object without touching the gripper, green object and the basket. \quad \quad \quad
    
    \vspace{0.4cm}
    
    \begin{wrapfigure}{L}{0.4\textwidth}
    \vspace{-0.5cm}
    \centering
    \includegraphics[width=0.4\textwidth]{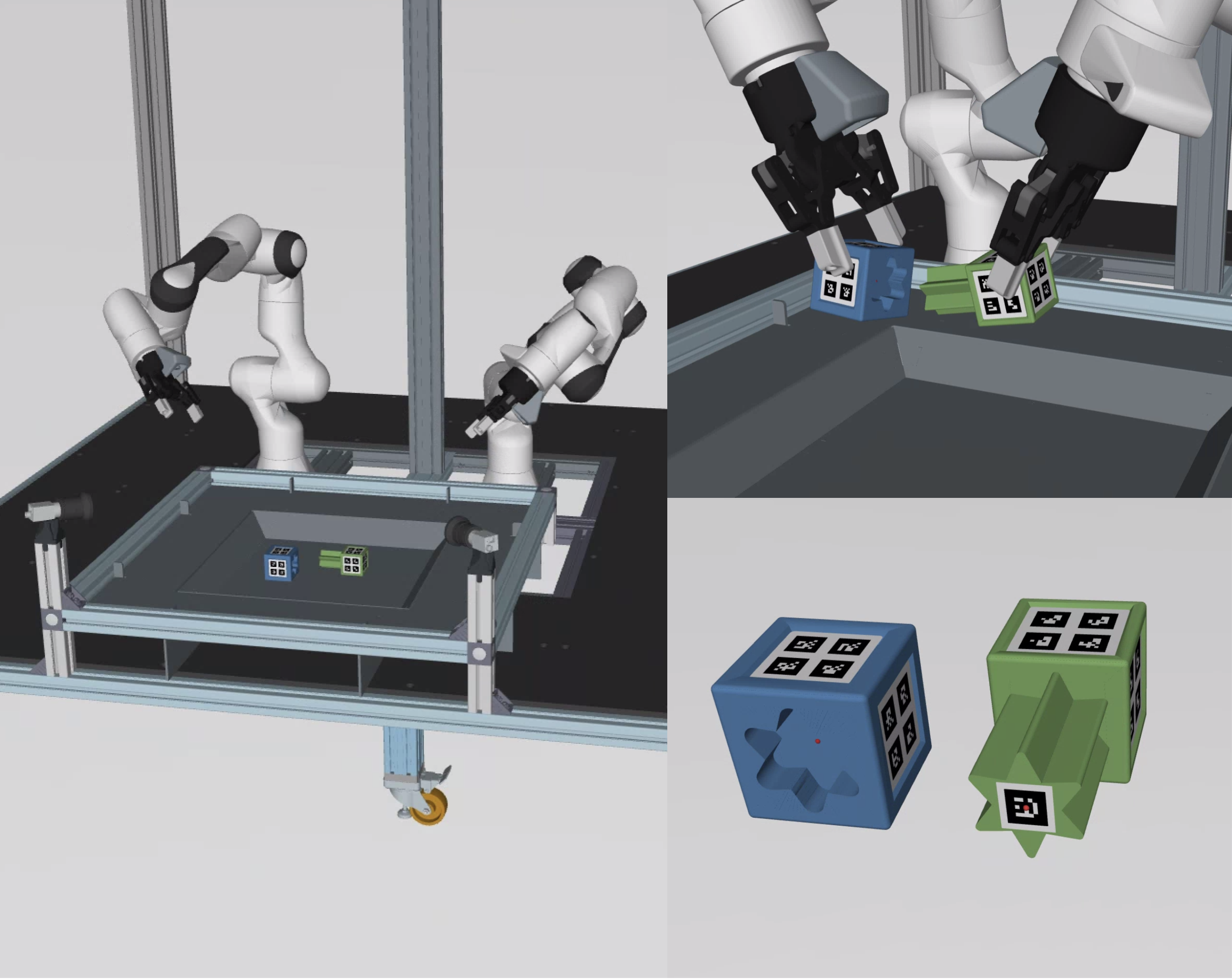}
    \vspace{-0.9cm}
    \end{wrapfigure}
    
    \paragraph{\textbf{Bimanual Insertion (State Size: 86, Action Size: 14)}}This environment features two 7 DOF arms that are each controlled using 6D Cartesian velocity controllers. Each arm has a 1 DOF gripper attached to it. The goal of the task is to insert the star shaped peg of the green object to star shaped hole of the blue object. The reward is zero when either of the objects' height from the basket is less than 0.15 meters, otherwise the reward is a function of the norm of the distance between the tip of the peg and the hole $dx = \left\Vert x_{peg} - x_{hole} \right\Vert_2$, where $r(dx) = \exp{(-(\alpha * dx)^2})$ with $\alpha=40$. If $r(dx) > 0.95$ it is rounded up to 1. 
    
    \vspace{0.3cm}
    
    \begin{wrapfigure}{L}{0.4\textwidth}
    \vspace{-0.5cm}
    \centering
    \includegraphics[width=0.4\textwidth]{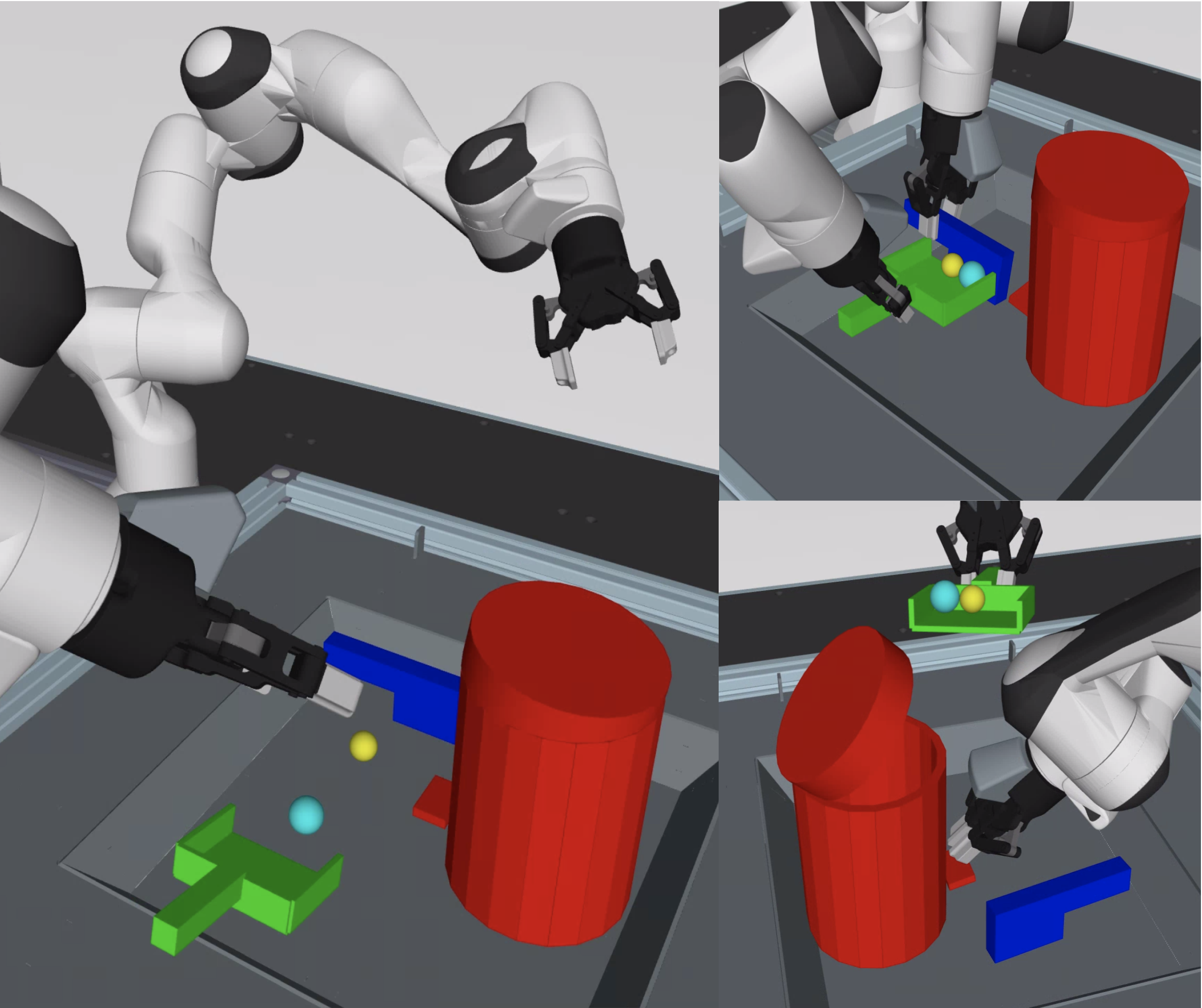}
    \vspace{-1.0cm}
    \end{wrapfigure}
    
    \paragraph{\textbf{Bimanual Cleanup (State Size: 129, Action Size: 14)}}This environment features two 7 DOF arms that are each controlled using 6D Cartesian velocity controllers. Each arm has a 1 DOF gripper attached to it. The goal of the task is to place the two balls inside the trashcan. The reward is 0.5 if either one of the balls are inside the trashcan and 1 if both balls are in the trashcan. Although not specified in the reward function, the desired behavior is for the robot to use both the brush and the dustpan to collect the balls and use the pedal to open the lid of the trashcan then pour in the balls in the trashcan. \quad  \quad \quad \quad \quad \quad  \quad  \quad \quad \quad \quad \quad \quad  \quad \quad \quad \quad \quad \quad  \quad 

    \vspace{0.35cm}
    
    \begin{wrapfigure}{L}{0.4\textwidth}
    \vspace{-0.5cm}
    \centering
    \includegraphics[width=0.4\textwidth]{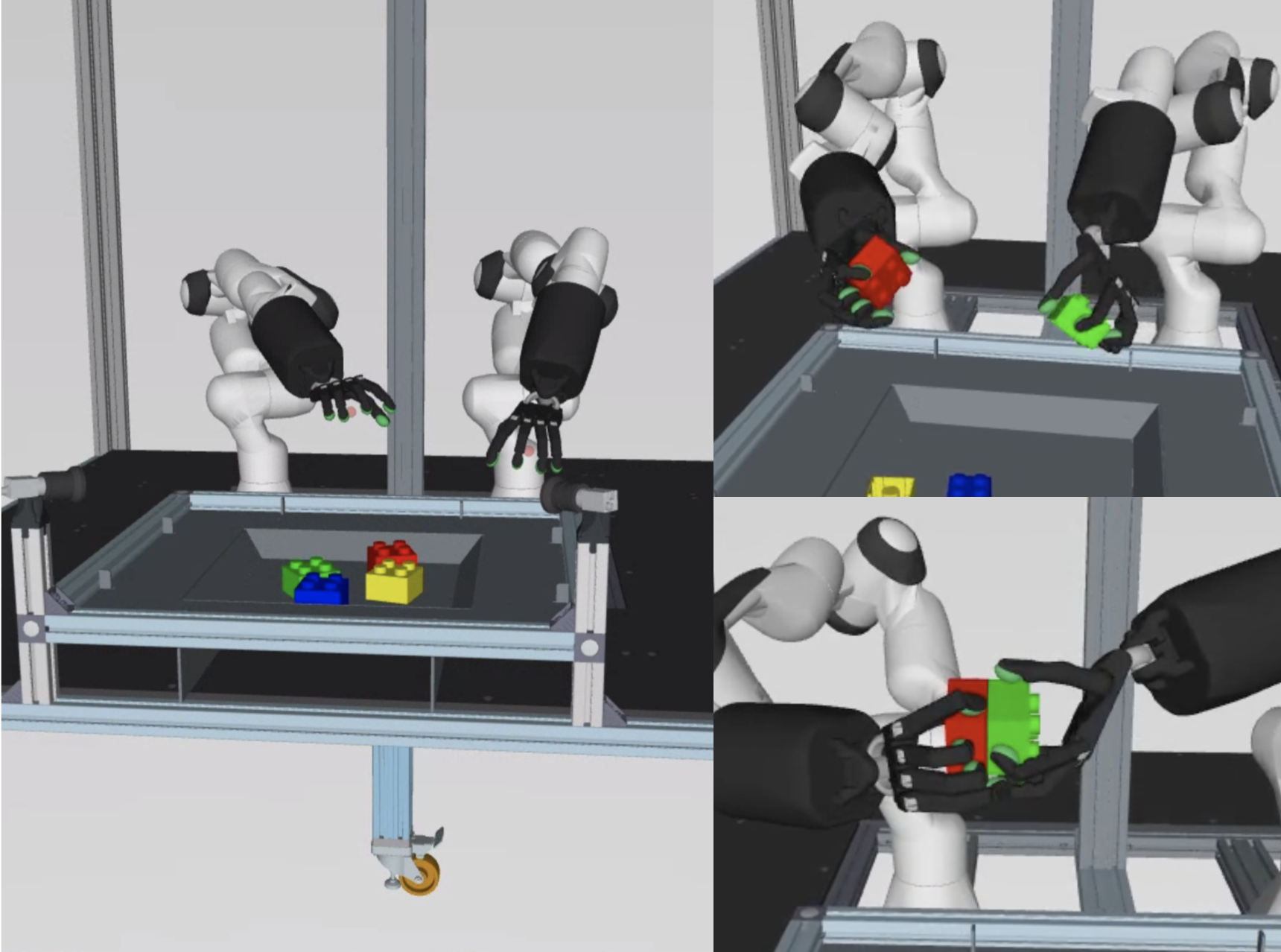}
    \vspace{-1.0cm}
    \end{wrapfigure}
    
    \paragraph{\textbf{Bimanual Shadow (State Size: 176, Action Size: 52)}}This environment features two 7 DOF arms that are each controlled using 6D Cartesian velocity controllers. Each arm has a 20 DOA/24 DOF hand attached to it. The goal of the task is to assemble the red and the green block together. The reward is zero when either of the blocks' height from the basket is less than 0.15 meters, otherwise the reward function is shaped to minimize the distance between the top of the red block to the bottom of the green block using the reward tolerance function from \citet{dm_control}.
    
    \vspace{0.5cm}

    All of the tasks use the MuJoCo physics state \citep{mujoco} as the observation. Each of the tasks have been chosen to probe certain types of challenges that are widely discussed in the robotics research field. The single arm stacking task is not only a widely used benchmark for robot learning research, but it also provides a challenging RL task with its simple sparse reward and wide initial state distribution. On the other hand, the bimanual insertion task aims to focus on precision, using two controllable arms, similar to how humans would perform the insertion task using their two arms. Hence the bimanual insertion task has a small stochastic initial state distribution with the peg and hole placed on right and left side of the basket respectively. The bimanual cleanup task focuses on high dimensional state space and tool use, hence it also has a small stochastic initial state distribution. The bimanual shadow task aims to bring  different types of complexity into a single task. It has a wide stochastic initial state distribution with the LEGO blocks starting anywhere on the basket with random orientations. The goal of assembling the LEGO blocks requires significant amount of precision and it has the largest state-action dimension with its human-like Shadow Hands.

\subsection{Theoretical Analysis of Relative Entropy Q-Learning}
\label{sec:req_theory}
        
     \paragraph{TD(0) Operator as a special case of REQ} It is easy to see that REQ performs policy evaluation for $\piprior$ when $\epsilon = 0$, as the only valid solution (up to differences which occur with zero probability) to the constrained optimization problem in \autoref{eq:constrained_kl} is $\pi = \piprior$. This is due to the Gibbs' inequality, where $\KL(P \| Q) = 0$ if and only if $P=Q$ almost everywhere. 
    
    \paragraph{Q-Learning  as a special case of REQ} Let us consider an MDP with a discrete action space with $\piprior$ having probability mass everywhere; $\piprior(a | s) > 0 \quad \forall a$. As $\epsilon \rightarrow \infty$ the solution to the constrained optimization problem in \autoref{eq:constrained_kl} approaches the greedy policy because the KL constraint is effectively removed, in which case the REQ operator is similar to that of Q-learning or DQN. It is worth noting that even when $\pi$ is $\argmax_a Q(s.a)$, the KL divergence is still finite due to the asymmetry of KL divergence, hence, as $\epsilon \rightarrow \infty$, $\pi$ can exactly represent $\argmax_a Q(s, a)$. In practice, when applied to a continuous action space, the maximization over action samples approximates the maximization over the action space.

    \paragraph{Policy Iteration Perspective of REQ} One way to analyze the REQ algorithm is as a simple policy iteration by considering the target Q-function to be fixed during the policy evaluation step. Specifically, the optimization in \autoref{eq:constrained_kl} is with respect to the target Q-function and hence the resulting policy is not conditioned on the current Q-function. In addition to the target Q-function for policy $\pi$, one can decide to use an additional target Q-function for the next state value estimate \citep{mnih2015humanlevel}, although we find that in practice we are able to use the same target Q-function for both purposes, as described in \autoref{eq:iw_req}. With the fixed Q-function for policy $\pi$, the contraction of the policy evaluation is obvious as its a simple TD(0) operator with respect to $\pi$. 
    
    When target Q-function is not used, then it bears strong similarity with the softmax Bellman operator discussed by \citet{littman_thesis} and \citet{softmax_operator}. With the main difference being that the temperature parameter $\eta_s$ in the REQ case is optimized adaptively to satisfy the KL constraint with respect to a prior policy $\piprior$ whereas the softmax Bellman operator uses a constant temperature value. It is worth noting that the softmax transform is not a non-expansion as shown by \citet{littman_thesis}. Surprisingly, despite its property, the softmax Bellman operator has been highly effectively shown by \citet{softmax_operator} as well as our own empirical results. In practice, we find that using a fixed epsilon with target networks is sufficient for good performance.
    
\subsection{KL constrained RL objective (closely following \citet{rerpi})}
\label{sec:reps_derivation}
    Let us consider the following constrained optimization problem for a given $s \in \mathcal{S}$, analogous to \autoref{eq:constrained_kl} in the main text but considering the optimization problem $\forall s \in \mathcal{S}$ given $\piprior$ and $\epsilon$:
    \begin{equation}
    \begin{aligned}
        \max_\pi & \int_\mathcal{A} \pi(a | s) Q^\pi(a, s) \diff a \\
        \textrm{s.t.} \quad & \KL(\pi(\cdot | s) \| \piprior(\cdot | s)) \leq \epsilon, \\
        & \int_\mathcal{A} \pi(a|s) \diff a = 1
        \label{eq:constrained_kl_appendix} \\
    \end{aligned}
    \end{equation}
    where the Lagrangian is,
    \begin{equation}
    \begin{aligned}
        L(\pi, \eta_s, \lambda_s) = &\int_\mathcal{A} \pi(a | s) Q^\pi(a, s) \diff a  + \\ & \eta_s \Bigg(\epsilon - \int_\mathcal{A}\pi(a|s) \log \frac{\pi(a|s)}{\piprior(a|s)} \diff a\Bigg) + \lambda_s \Bigg(1 -  \int_\mathcal{A}\pi(a|s) \diff a\Bigg)
        \label{eq:lagrangian} \\
    \end{aligned}
    \end{equation}
    Differentiating with respect to $\pi$ yields, 
    \begin{equation}
       \frac{\partial L}{\partial \pi} = Q^\pi(a, s) - \eta_s \log \pi(a|s)+ \eta_s \log \piprior(a|s) - (\eta_s - \lambda_s)
        \label{eq:lagrangian_pi} 
    \end{equation}
    where setting it to zero and rearranging terms to solve for $\pi$ we get,
    \begin{equation}
       \pi(a|s) = \piprior \exp\Bigg( \frac{Q^\pi(a, s)}{\eta_s}\Bigg) \exp\Bigg( -\frac{\eta_s - \lambda_s}{\eta_s}\Bigg)
        \label{eq:exp_q_0_appendix} 
    \end{equation}
    It is easy to see that when $\int_\mathcal{A} \pi(a|s) \diff a = 1$ is satisfied, $\pi(a|s)$ is a valid probability distribution, hence, the last term in \autoref{eq:exp_q_0_appendix} is the normalizing constant for $\pi$, and we can write,
    \begin{equation}
    \begin{aligned}
       \exp\Bigg(\frac{\eta_s - \lambda_s}{\eta_s}\Bigg) = \int_\mathcal{A }\piprior \exp\Bigg( \frac{Q^\pi(a, s)}{\eta_s}\Bigg) \diff a \\
       \frac{\eta_s - \lambda_s}{\eta_s} = \log \Bigg(\int_\mathcal{A }\piprior \exp\Bigg( \frac{Q^\pi(a, s)}{\eta_s}\Bigg) \diff a \Bigg)\\
        \label{eq:exp_q_norm} 
    \end{aligned}
    \end{equation}
    The dual function $g(\eta_s)$ can be obtained by inserting above into the Lagrangian,
    \begin{equation}
    \begin{aligned}
        L(\pi, \eta_s, \lambda_s) = &\int_\mathcal{A} \pi(a | s) Q^\pi(a, s) \diff a  \\ & - \eta_s \int_\mathcal{A} \pi(a|s) \Bigg(\frac{Q^\pi(a, s)}{\eta_s} + \log \piprior(a|s) - \frac{\eta_s - \lambda_s}{\eta_s} \Bigg) \diff a + \eta_s \epsilon \\
        & + \eta_s \int_\mathcal{A} \pi(a|s) \log \piprior(a|s) \diff a + \lambda_s \Bigg(1 -  \int_\mathcal{A}\pi(a|s) \diff a\Bigg)
        \label{eq:lagrangian_0} \\
    \end{aligned}
    \end{equation}
    where simplifying yields,
    \begin{equation}
        L(\pi, \eta_s, \lambda_s) = \eta_s \epsilon + \eta_s \frac{\eta_s - \lambda_s}{\eta_s} 
        \label{eq:lagrangian_eta} \\
    \end{equation}
    and plugging in \autoref{eq:exp_q_norm} gives the dual function for the Lagrangian multiplier $\eta_s$,
    \begin{equation}
        g(\eta_s) = \eta_s \epsilon + \eta_s \log \Bigg(\int_\mathcal{A }\piprior \exp\Bigg( \frac{Q^\pi(a, s)}{\eta_s}\Bigg) \diff a \Bigg).
        \label{eq:dual} \\
    \end{equation}
    We optimize the dual function $g(\eta_s)$ using multiple steps of gradient decent. The initial value of the temperature is set to the mean of the standard deviation over the number of action samples of the Q values,
    \begin{equation}
        \eta_0 = \frac{1}{N}\sum_n \sigma_M(Q(s,a_{nm})) \\
    \end{equation}
    where $N$ is size of the mini batch, $n$ its and the $M$ is the number of action samples, $m$ its index. Additionally the gradient is scaled using the current value of the temperature, to ensure reliable convergence for arbitrary set of Q values. 
    
\subsection{Waypoint Tracking Controllers as Suboptimal Expert}
\label{sec:internal_agent}

    The feedback linear controllers described in \autoref{sec:rlfse} are concatenated sequentially (optionally with learnt primitives) to construct suboptimal experts. A full listing of our procedure is presented in Algorithm \ref{alg:internal_agent}.

    \begin{algorithm}[H]
    \small
    \caption{Waypoint Tracking Controllers as Suboptimal Expert}\label{alg:internal_agent}    \begin{algorithmic}
    \STATE \textbf{Input:} State $s$, $N$ sequence of motions. Initialize index for motion $i=0$, Motion is a tuple of (action $\bar{a}$, $M$ target frames $\mathcal{F}$, optional learned controller $\psi_\pi$, optional jump function $J$).
    \STATE $a_i \leftarrow \bar{a}$
    \WHILE{$j \leq M$}
        \STATE set position velocity action $a_{i, j, p} \leftarrow \psi_{p}(s, \mathcal{F}_j)$
        \STATE set orientation velocity action $a_{i, j, o} \leftarrow \psi_{o}(s, \mathcal{F}_j)$
    \ENDWHILE
    \IF{$\psi_\pi(s)$}
        \STATE $a_i \leftarrow \psi_\pi(s)$
    \ENDIF
    \IF{\textit{timeout}}
        \STATE $i \leftarrow i+1$
    \ENDIF
    \IF{$J$}
        \STATE $i \leftarrow J(s)$
    \ENDIF
    \STATE \textbf{return} $a_i$
    \end{algorithmic}
    \end{algorithm}
    
    \paragraph{\textbf{Single Arm Stacking}}While using action $\bar{a}$ to specify the gripper to be open, the end-effector of the single arm is taken above the frame of the red object and lowered. The action $\bar{a}$ is updated to close the gripper. The end-effector is taken above the blue object. The action $\bar{a}$ is then updated to open the gripper.
    
    \paragraph{\textbf{Bimanual Insertion}}Both of the grippers are set to open in $\bar{a}$ and the two end-effectors are taken to the peg and the hole object. Both of the grippers are set to close in $\bar{a}$. The end-effectors are lifted up and the peg and the hole are brought close together.
    
    \paragraph{\textbf{Bimanual Cleanup}}Both of the grippers are set to open in $\bar{a}$ and the two end-effectors are taken to the brush and the dustpan object. Both of the grippers are set to close in $\bar{a}$. The end-effectors brought close together while staying close to the basket in attempt to lift up the two balls. The end-effectors lifted. The left end-effector is taken to pedal of the trashcan and lowered. The right end-effector is taken above the trashcan and the dustpan is tilted down towards the opening trashcan.
    
    \paragraph{\textbf{Bimanual Shadow LEGO}}The bimanual Shadow Hand LEGO stacking task is the only task which uses learned primitives. Two learned primitives are are trained using standard off-policy RL with REQ. The first primitive is trained to orient the green LEGO to an upwards facing position with the left hand. The second primitive is trained to orient the red LEGO block to a downwards facing position with the right hand. In both cases the block must remain in the basket. The orient tasks uses a shaped reward as a function of the orientation error defined in \autoref{sec:rlfse} with respect to the closest target orientation (taking into account the four symmetric orientations of the LEGO blocks). At the start of the task, both of the Shadow Hands are is set to open in $\bar{a}$ and the right hand is taken above the red LEGO block. The learned primitive for the right hand executes in order to orient the red LEGO block to a downwards facing position. The right hand is updated to close in $\bar{a}$ and the right hand is lifted. If the red LEGO block is not lifted, it retries using the jump function $J$. The green LEGO block is lifted in a similar way using the left hand. The red and the green LEGO blocks are then aligned and brought closer together.

\subsection{Intertwining Exploration}
\label{sec:intertwining}
    \begin{algorithm}[H]
    \small
    \caption{Intertwining Exploration}\label{alg:intertwining}    \begin{algorithmic}
    \STATE \textbf{Input:} policy $\pi$, suboptimal policy $\psi$, number of episodes $N$, $\lambda_\psi$ and $\lambda_{intertwine}$.
    \WHILE{$i \leq N$}
    \STATE Reset the environment
    \STATE \textit{should\_intertwine} = $uniform(0,1) < \lambda_{intertwine}$
        \WHILE{\textit{episode not terminated}}
        \IF{\textit{should\_interwine}}
            \STATE \textit{act\_with\_$\psi$} = $uniform(0,1) < \lambda_\psi$
        \ENDIF
        
        $a = \begin{cases} a \sim \pi(a|s) \text{ if }\textit{act\_with\_$\psi$} \\ a \sim \psi(a|s)  \quad \text{else}, \end{cases}$
        \STATE Act in the environment with $a$
        \ENDWHILE
    \ENDWHILE
    \end{algorithmic}
    \end{algorithm}

\subsection{RLfD, BC and DAgger Results}
\label{sec:rlfd_rlfse}

\begin{figure}[h]
    \centering
    \includegraphics[width=1.0\linewidth]{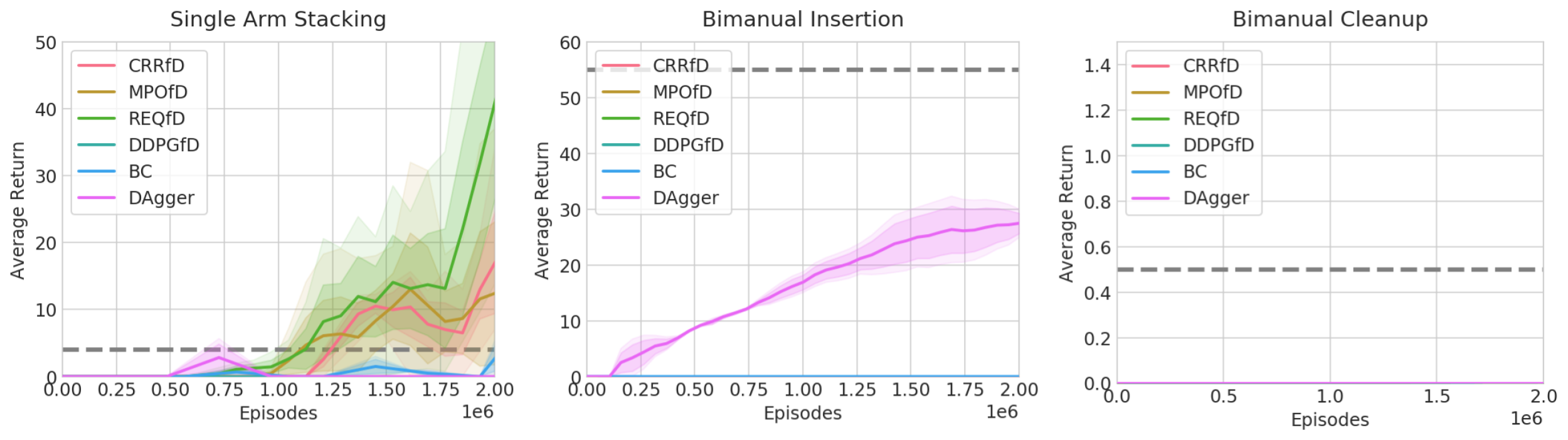}  
    \caption{
    RLfD results on the tasks shown in \autoref{fig:rlfse_envs}. The dotted line represents the performance of the suboptimal expert. Note that the DAgger refers to using the same intertwining exploration for RLfSE but optimizing the negative log-likelihood of the actions from the suboptimal expert.
     }
    \label{fig:rlfd}
    \end{figure}

    The results in \autoref{fig:rlfd} clearly shows that RLfD, BC and DAgger struggling to learn the manipulation tasks especially for the bimanual tasks. Overall, the results for all the algorithms in \autoref{fig:rlfd} is significantly less than the results of our REQfSE algorithm in \autoref{fig:rlfse_rlfd_plot}. It is worth noting that the we only use MLP networks for these three tasks and that the suboptimal expert has an internal counter, which could explain the ineffectiveness of pure imitation learning algorithms, BC and DAgger.
    
\subsection{Offline Reinforcement Learning Details}
\label{sec:offline}
As described in the main text, the offline RL results use the same network architecture, same dataset and the evaluation criteria as \citet{crr} to ensure a controlled and fair experiments.

\newpage

\subsection{Ablation Studies}
\label{sec:ablation}

    \begin{figure}[!htb]
    \centering
    \includegraphics[width=1.0\linewidth]{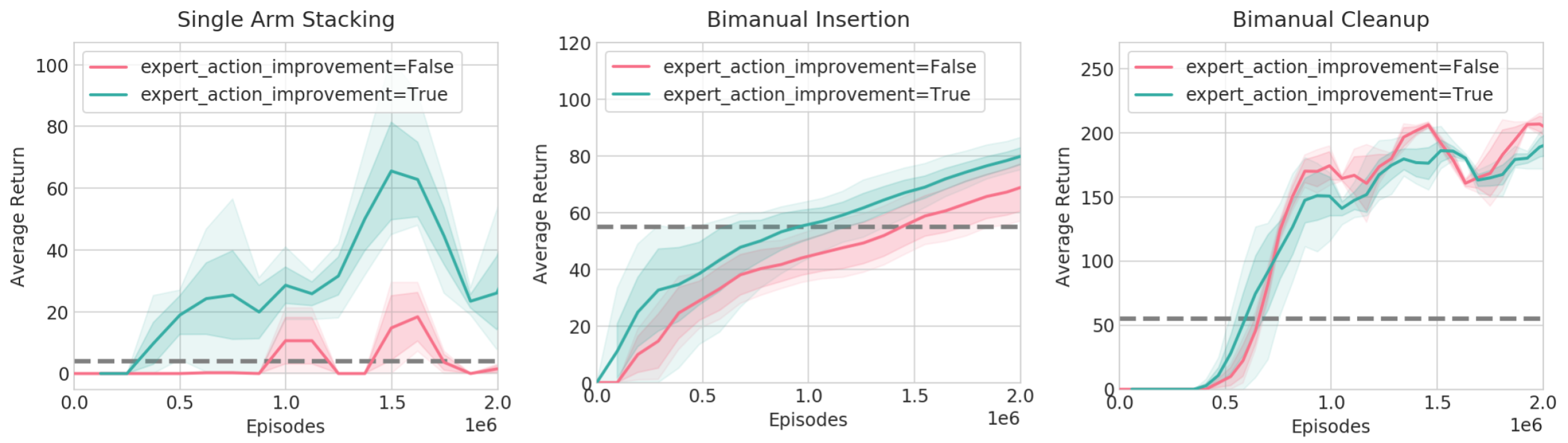} 
    \caption{
    Ablation on using the \textit{expert\_action\_improvement} described in \autoref{eq:policy_improvement}.
     }
    \label{fig:expert_action_improvemnt}
    \end{figure}

    \begin{figure}[!htb]
    \centering
    \includegraphics[width=1.0\linewidth]{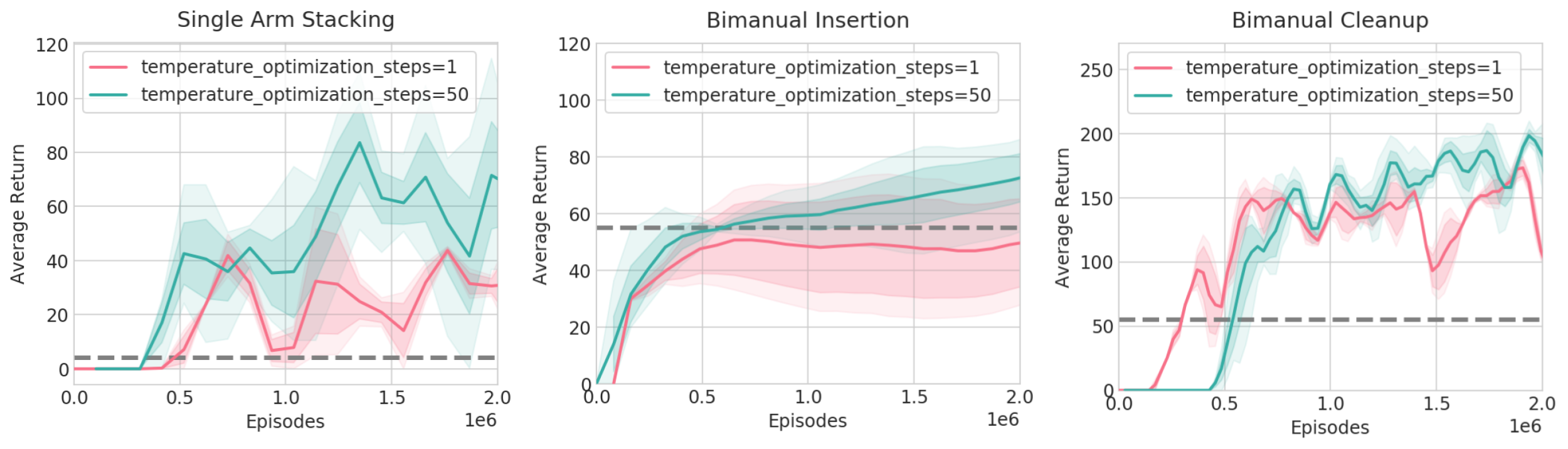} 
    \caption{
    Ablation on number of gradient steps for the temperature described in \autoref{sec:req}.
     }
    \label{fig:reqfse_num_iter}
    \end{figure}

    \begin{figure}[!htb]
    \centering
    \includegraphics[width=1.0\linewidth]{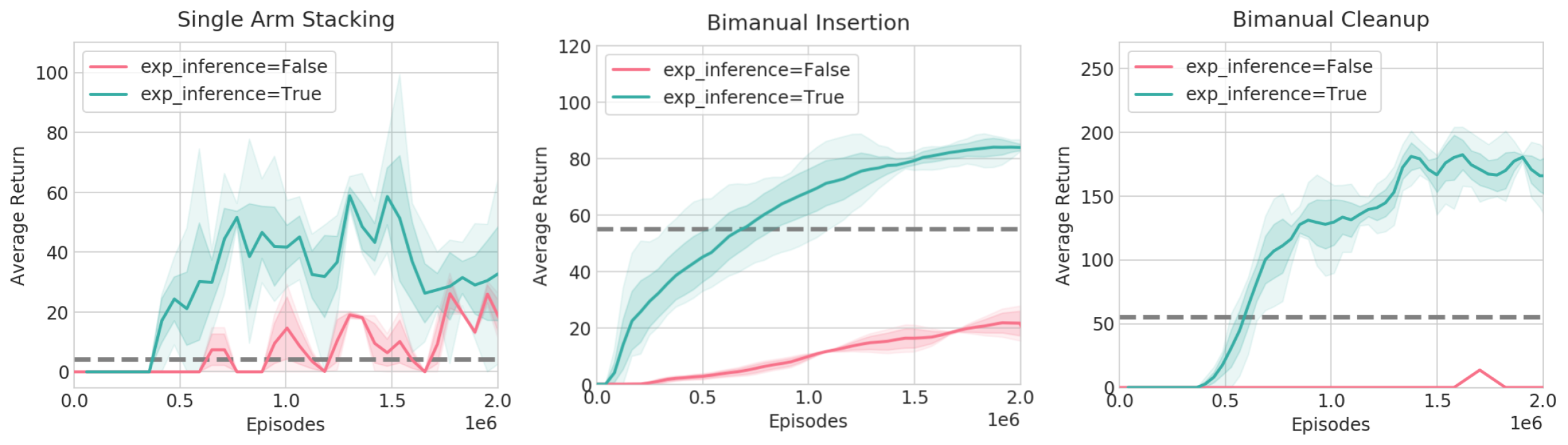} 
    \caption{
    Ablation of using the exponential transform during inference or using the prior policy for acting described in \autoref{sec:req}.
     }
    \label{fig:req_exp_inf}
    \end{figure}

    \begin{figure}[!htb]
    \centering
    \includegraphics[width=1.0\linewidth]{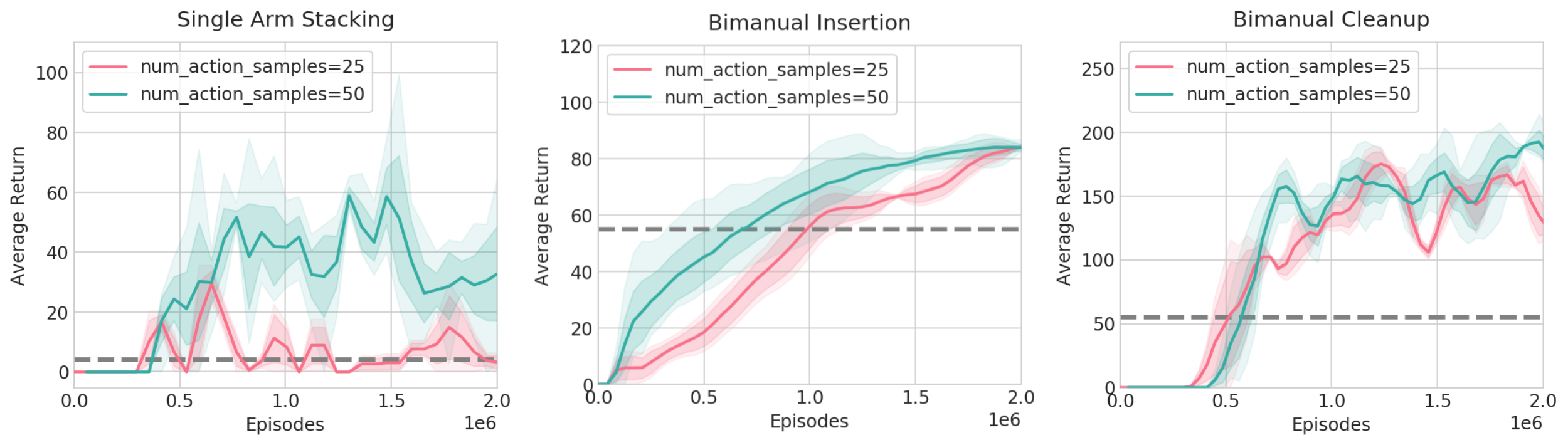} 
    \caption{
    Ablation on number of action samples for value estimate and importance sampling described in \autoref{sec:req}.
     }
    \label{fig:req_num_a_sample}
    \end{figure}

    \begin{figure}[!htb]
    \centering
    \includegraphics[width=1.0\linewidth]{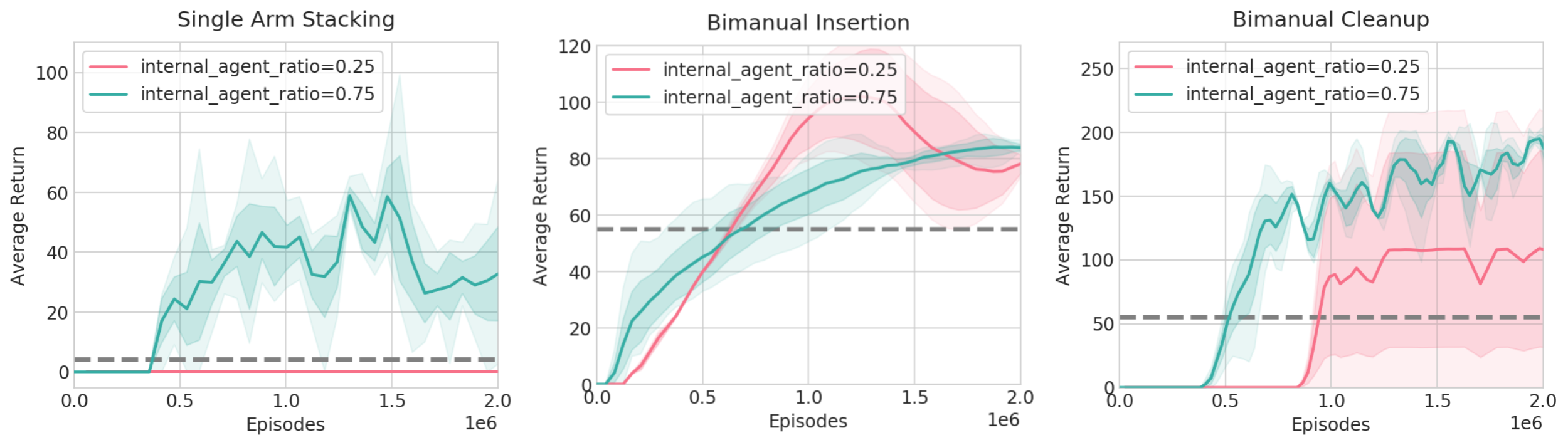} 
    \caption{
    Ablation on the probability of acting with the suboptimal expert described in \autoref{sec:req}.
     }
    \label{fig:req_internal_agent}
    \end{figure}

    \begin{figure}[!htb]
    \centering
    \includegraphics[width=1.0\linewidth]{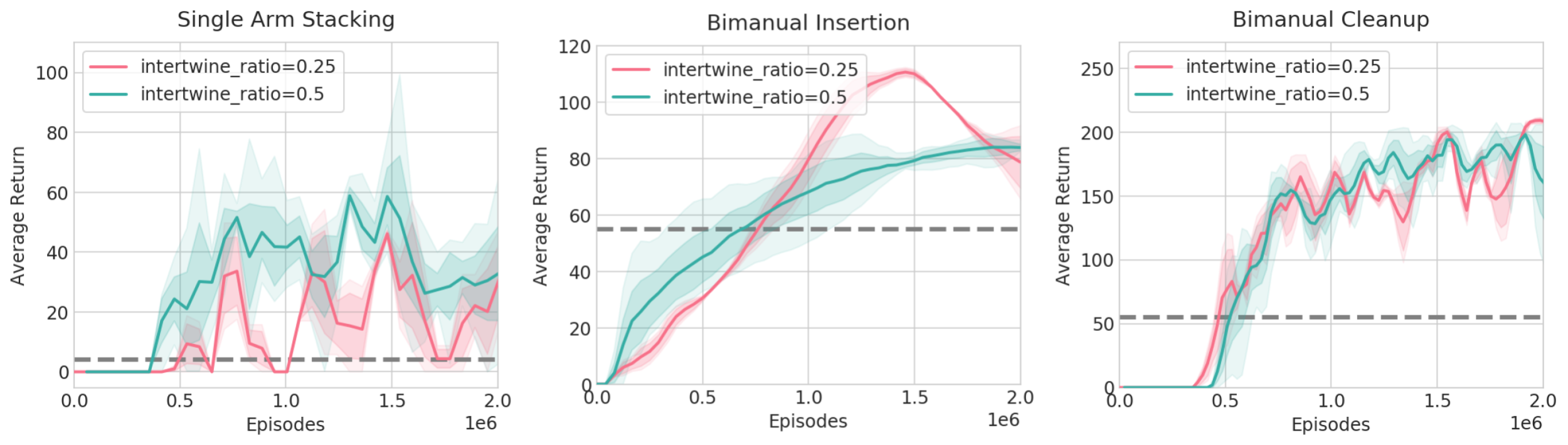} 
    \caption{
    Ablation on the intertwining exploration probability described in \autoref{sec:req}.
     }
    \label{fig:req_intertwine}
    \end{figure}
    
    \begin{figure}[!htb]
    \centering
    \includegraphics[width=1.0\linewidth]{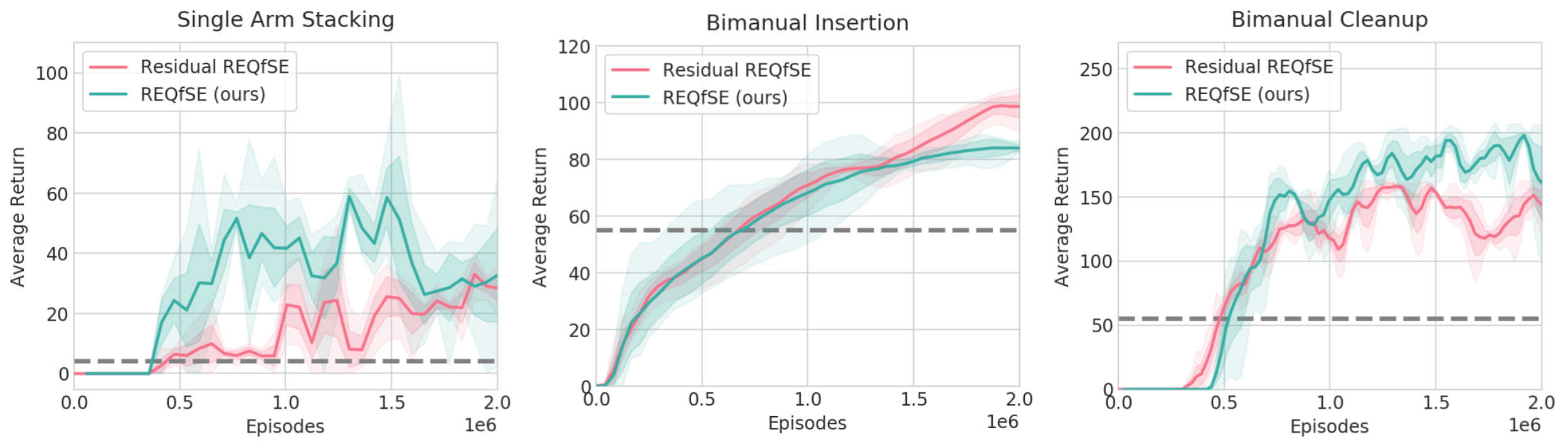} 
    \caption{
    Comparison to a setup similar to that of residual reinforcement learning \citep{residual_rl} for REQfSE, where the action of the suboptimal expert is provided as an input to agent.
     }
    \label{fig:req_residual}
    \end{figure}
    
    \begin{figure}[!htb]
    \centering
    \includegraphics[width=1.0\linewidth]{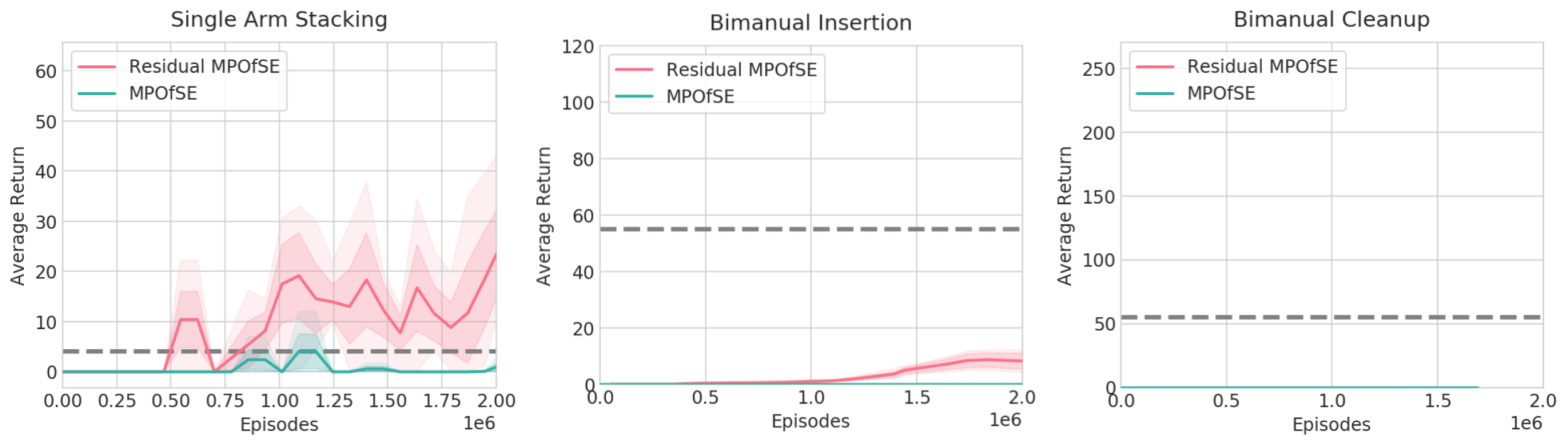} 
    \caption{
    Comparison to a setup similar to that of residual reinforcement learning \citep{residual_rl} for MPOfSE, where the action of the suboptimal expert is provided as an input to agent.
     }
    \label{fig:mpo_residual}
    \end{figure}

\subsection{Hyperparameters}
\label{sec:hyper}
Neural networks are assumed to be MLP unless stated otherwise.\footnote{The bimanual shadow task with REQfSE used slightly different hyperparameters with $\lambda_\psi = 0.5$ and using an LSTM instead of the middle MLP.}

\begin{table}[h]
 \small
\begin{center}
 \begin{tabular}{c||c} 
 Hyperparameters & CRR $|$ REQ \\
 \hline
 Policy net & 256-256-256 \\
 Q function net & 256-256-256 \\
 Discount factor ($\gamma$) & 0.99 \\
 Adam learning rate & 0.0003 \\
 Replay buffer size & 1000000 \\
 Target network update period & 20 \\
 Batch size & 512 \\
 Unroll length & 2 \\
 Activation function & elu\\
 Layer norm on first layer & Yes\\
 Min variance & 0.00001\\
 Max variance & unbounded\\
 Number of action samples & 20\\
 REQ epsilon ($\epsilon$) & 0.75\\
 Number of gradient decent steps for $\eta_s$ & 20\\
 Epsilon mean ($\epsilon_\mu$) & 0.01\\
 Epsilon covariance ($\epsilon_\mu$) & 0.00001\\
 Policy evaluation method & TD(0) Operator $|$ REQ Operator \\
\end{tabular}
\end{center}
\caption{Hyperparameters for CRR and REQ}
\end{table}

\begin{table}[h]
 \small
\begin{center}
 \begin{tabular}{c||c} 
 Hyperparameters & DDPGfD and DDPGfSE \\
 \hline
 Policy net & 256-256-256 \\
 Q function net & 256-256-256 \\
 Discount factor ($\gamma$) & 0.99 \\
 Adam learning rate & 0.0001 \\
 Replay buffer size & 1000000 \\
 Target network update period & 200 \\
 Batch size & 128 \\
 Unroll length & 32 \\
 Activation function & elu\\
 Layer norm on first layer & Yes\\
 Action noise standard deviation & 0.3\\
\end{tabular}
\end{center}
\caption{Hyperparameters for DDPGfD and DDPGfSE}
\end{table}

\begin{table}[h]
\small
\begin{center}
 \begin{tabular}{c||c} 
 Hyperparameters & MPOfD and MPOfSE \\
 \hline
 Policy net & 256-256-256 \\
 Q function net & 256-256-256 \\
 Discount factor ($\gamma$) & 0.99 \\
 Adam learning rate & 0.0001 \\
 Replay buffer size & 1000000 \\
 Target network update period & 200 \\
 Batch size & 128 \\
 Unroll length & 32 \\
 Activation function & elu\\
 Layer norm on first layer & Yes\\
 Min variance & 0.00001\\
 Max variance & unbounded\\
 Number of action samples & 20\\
 MPO epsilon & 0.1\\
 Epsilon mean ($\epsilon_\mu$) & 0.001\\
 Epsilon covariance ($\epsilon_\mu$) & 0.00001\\

\end{tabular}
\end{center}
\caption{Hyperparameters for MPOfD and MPOfSE}
\end{table}

\begin{table}[h]
\small
\begin{center}
 \begin{tabular}{c||c} 
 Hyperparameters & CRRfSE and CRRfSE $|$ REQfD and REQfSE \\
 \hline
 Policy net & 256-256-256 \\
 Q function net & 256-256-256 \\
 Discount factor ($\gamma$) & 0.99 \\
 Adam learning rate & 0.0001 \\
 Replay buffer size & 1000000 \\
 Target network update period & 500 \\
 Batch size & 128 \\
 Unroll length & 32 \\
 Activation function & elu\\
 Layer norm on first layer & Yes\\
 Min variance & 0.00001\\
 Max variance & unbounded\\
 Number of action samples & 50\\
 REQ epsilon ($\epsilon$) & 0.75\\
 Number of gradient decent steps for $\eta_s$ & 50\\
 Epsilon mean ($\epsilon_\mu$) & 0.005\\
 Epsilon covariance ($\epsilon_\mu$) & 0.00001\\
 Policy evaluation method & TD(0) Operator $|$ REQ Operator \\
\end{tabular}
\end{center}
\caption{Hyperparameters for CRRfSE, CRRfSE, REQfD and REQfSE}
\end{table}

\begin{table}[h]
\small
\begin{center}
 \begin{tabular}{c||c||c} 
 Hyperparameters & RLfD & RLfSE \\
 \hline
 $\lambda_\psi$ & 0.25 & 0.75 \\
 $\lambda_{intertwine}$ & 0.0 & 0.5 \\
\end{tabular}
\end{center}
\caption{Hyperparameters for Intertwine  Exploration}
\end{table}


\end{document}